\documentclass[10pt,twocolumn,letterpaper]{article}

\usepackage[final]{cvpr}
\usepackage[pagebackref,breaklinks,colorlinks,allcolors=cvprblue]{hyperref}
\usepackage{array}

\usepackage{xcolor,colortbl}
\usepackage{booktabs}
\usepackage{pifont}
\usepackage{soul}
\usepackage[font=small,labelfont=bf,tableposition=top]{caption}

\usepackage{graphicx}
\usepackage{amsmath}
\usepackage{amssymb}
\usepackage{makecell}
\usepackage{subcaption}
\usepackage{caption}
\usepackage{multirow}
\usepackage{multicol}
\usepackage[dvipsnames]{xcolor}
\usepackage{soul}
\usepackage{comment}
\usepackage{bbding}
\usepackage{makecell}
\usepackage{xspace}
\usepackage{soul}

\definecolor{cvprblue}{rgb}{0.21,0.49,0.74}
\definecolor{emphasize}{RGB}{42,69,189}
\definecolor{linkcolor}{RGB}{231,68,149}

\newcommand{\ours}{DiET-GS\xspace}
\newcommand{\ourspp}{DiET-GS++\xspace}

\title{DiET-GS: Diffusion Prior and Event Stream-Assisted \\ Motion Deblurring 3D Gaussian Splatting}

\author{Seungjun Lee \qquad Gim Hee Lee\\
Department of Computer Science, National University of Singapore\\
{\tt\small seungjun.lee@u.nus.edu, gimhee.lee@nus.edu.sg} \\ \vspace{-5pt} \\ Project page: \href{https://diet-gs.github.io}{\textcolor{linkcolor}{DiET-GS.github.io}}}

\begin{document}

\twocolumn[{
\renewcommand\twocolumn[1][]{#1}
\maketitle
\vspace{-27pt}
\begin{center}
    \captionsetup{type=figure}
    \includegraphics[width=\textwidth]{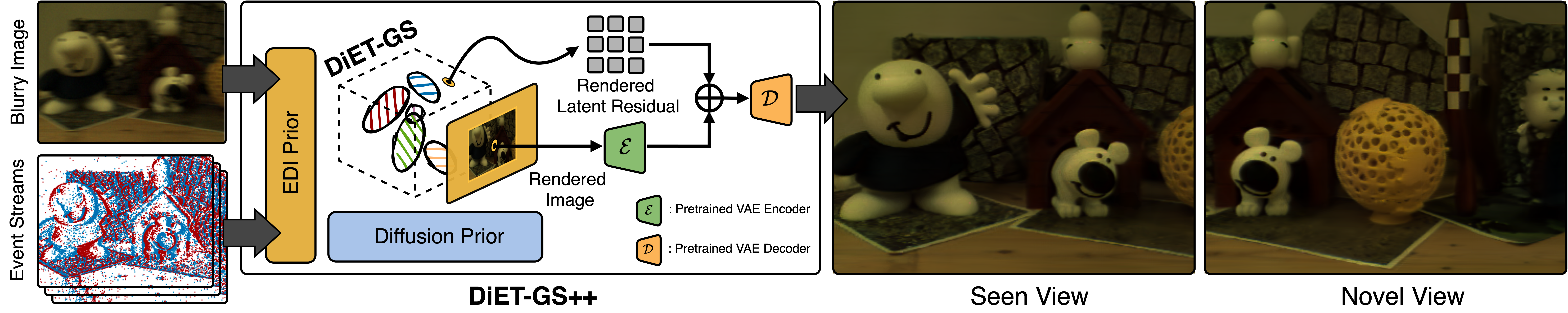}
    \caption{Given a set of blurry images and corresponding event streams, we propose a novel framework to construct deblurring 3DGS by jointly leveraging the EDI~\cite{pan2019bringing} formulation and a pretrained diffusion model as a prior. Our \ourspp enables high quality novel-view synthesis with recovering precise color and well-defined details from the blurry multi-view images.}
    \label{fig:teaser}
    \end{center}
}]

\begin{abstract}
\indent Reconstructing sharp 3D representations from blurry multi-view images is a long-standing problem in computer vision. Recent works attempt to enhance high-quality novel view synthesis from the motion blur by leveraging event-based cameras, benefiting from high dynamic range and microsecond temporal resolution. However, they often reach sub-optimal visual quality in either restoring inaccurate color or losing fine-grained details. In this paper, we present \ours, a diffusion prior and event stream-assisted motion deblurring 3DGS. Our framework effectively leverages blur-free event streams and diffusion prior in a two-stage training strategy.
Specifically, we introduce the novel framework to constrain 3DGS with event double integral, achieving both accurate color and well-defined details. Additionally, we propose a simple technique to leverage diffusion prior to further enhance the edge details. Qualitative and quantitative results on both synthetic and real-world data demonstrate that our \ours is capable of producing better quality of novel views compared to the existing baselines. 

\vspace{-20pt}
\end{abstract}    
\section{Introduction}
\label{sec:intro}

Novel view synthesis plays an important role in various vision applications such as scene understanding~\cite{kundu2022panoptic, liu2023unsupervised, xie2021fig}, virtual reality~\cite{xu2023vr, laviola2008bringing}, image processing~\cite{huang2022hdr, ma2022deblur, mildenhall2022nerf}, \etc. To this end, Neural Radiance Fields (NeRF)~\cite{mildenhall2021nerf} achieves notable success in generating high-quality novel views by reconstructing implicit 3D representations with a deep neural network. However, it falls short of real-time rendering and efficient training, limiting its application in real-world scenarios. Recently, 3D Gaussian Splatting (3DGS)~\cite{kerbl20233d} has emerged as an efficient alternative to NeRF by representing scenes with explicit 3D Gaussian primitives. The explicit representation of 3DGS serves as a lightweight replacement of the implicit neural representations used in NeRF, achieving better training and rendering efficiency as well as visual quality of novel views.

To enable high-quality 3D reconstructions, both NeRF and 3DGS rely on multi-view images that are perfectly captured and free from any artifact. However, this preliminary condition is often unavailable in the real world. For example, a camera needs prolonged exposure time in a low light environment to allow enough light to reach the sensor for image formation. The camera has to remain absolutely still during this lengthy exposure time. Any camera motion during the capture leads to undesired motion blur. To circumvent this issue, a line of works~\cite{peng2022pdrf, lee2023dp, ma2022deblur, wang2023bad, zhao2025bad} has attempted to recover sharp 3D representations from blurry multi-view images. Despite the promising potential, it is non-trivial to restore fine-grained details from blurry images alone and thus often leading to sub-optimal visual quality.

Several recent works have shown the efficacy of event-based cameras, significantly improving motion deblurring in images captured from standard frame-based cameras. Event sensors enable blur-free measurements of brightness changes, benefiting from higher dynamic range and temporal resolution compared to standard cameras. Motivated by this distinct competency, several recent works have explored the potential of recovering sharp 3D representations from event streams. Earlier works~\cite{rudnev2023eventnerf, hwang2023ev, bhattacharya2024evdnerf} focus on utilizing solely event-based data, lacking the capacity to preserve color information. E-NeRF~\cite{klenk2023nerf} combines blurry images into the framework as direct color supervision for 3D NeRF. Nonetheless, the estimated color exhibits blur around the edges since it does not account for the blur formation. E$^2$NeRF and following works~\cite{cannici2024mitigating, li2025benerf, yu2024evagaussians, qi2024deblurring} explicitly model the blur formation process to further enhance the color and edge details. However, most of these existing works still rely on blurry images alone to recover accurate color, often resulting in unwanted color artifacts. To supplement color guidance, Ev-DeblurNeRF~\cite{cannici2024mitigating} proposes to exploit the explicit relationship of Event Double Integral (EDI) between blurry images and event streams.

In this paper, we propose \textcolor{emphasize}{\textbf{\ours}}, a \underline{Di}ffusion prior and \underline{E}ven\underline{T} stream-assisted motion deblurring 3D\underline{GS}. As illustrated in Fig.~\ref{fig:teaser}, our framework comprises two main stages: \ours and \ourspp. Our \textcolor{emphasize}{\textbf{Stage 1: \ours}} first optimizes the deblurring 3DGS by jointly leveraging the real-captured event streams and the prior knowledge of a pretrained diffusion model. To restore both accurate color and well-defined details, we introduce a novel framework that uses the EDI prior to achieve 1) \textit{fine-grained details}, 2) \textit{accurate color}, and 3) \textit{regularization}. Specifically, in addition to the EDI color guidance proposed by~\cite{cannici2024mitigating}, we propose further constraints to recover fine-grained details by modeling EDI in the brightness domain through a learnable camera response function. This learnable approach naturally considers the potential variation between RGB values and pixel intensity, leading to better real-world adaptation and thus effectively recovering intricate details. The EDI constraints from both RGB space~\cite{cannici2024mitigating} and brightness domain enable mutual compensation between color fidelity and fine-grained details, resulting in optimal visual quality. Additionally, we derive a regularization term from the EDI prior to further facilitate the optimization by ensuring the cycle consistency among the objective terms.

To achieve more natural image refinement, we further incorporate diffusion prior in \ours using the Renoised Score Distillation (RSD) proposed by~\cite{lee2024disr}. Nonetheless, we empirically find that jointly leveraging both priors from real-captured data and a pretrained diffusion model often weakens the full effect of diffusion prior. Consequently, our \textcolor{emphasize}{\textbf{Stage 2: \ourspp}} is further introduced to maximize the effect of diffusion guidance by adding extra learnable parameters to each 3D Gaussian in the pretrained \ours. Unlike~\cite{lee2024disr}, \ourspp directly renders latent residual from the 3D Gaussians, resulting in a simpler framework to leverage RSD optimization. Finally, resulting images from \ours are further refined by solely relying on diffusion prior while edge details are effectively enhanced.

Our main contributions are summarized as follows:
\begin{itemize}
    \item 
    A novel framework to construct deblurring 3DGS by jointly leveraging event streams and the prior knowledge of a pretrained diffusion model.
    \item
    A two-stage train38ing strategy to effectively utilize real-captured data and diffusion prior together. Once optimized, our method is capable of recovering well-defined details with accurate color from the input blurry images.
    \item Qualitative and quantitative results show that our framework significantly surpasses the existing baselines, achieving the best visual quality.
\end{itemize}

\vspace{-5pt}
\section{Related Works}
\label{sec:related_works}
\noindent \textbf{Event-based image deblurring.} Recently, event-based cameras have gained significant popularity due to their high dynamic range and microseconds temporal resolution. Several methods have tried to leverage these distinct features to tackle image deblurring. Most notably, event-based double integral (EDI)~\cite{pan2019bringing} explicitly models the relationship between event streams and a blurry image jointly captured during the fixed exposure time. Subsequent works follow EDI by fusing the events and RGB frames~\cite{haoyu2020learning, sun2022event, sun2023event, xu2021motion, zhang2020hybrid} or using learning-based approaches~\cite{jiang2020learning, wang2020event} to further improve visual quality. 
Another line of works exploits event cameras to recover sharp 3D representations such as NeRF and 3DGS from blurry multi-view images. Earlier works~\cite{hwang2023ev, rudnev2023eventnerf, low2023robust, klenk2023nerf} utilize an event generation model~\cite{gallego2020event} to enable sharp novel view synthesis from a fast-moving camera while E-NeRF~\cite{klenk2023nerf} combines RGB frames to further refine the color. Recent works~\cite{qi2023e2nerf, cannici2024mitigating, qi2024deblurring, yu2024evagaussians, li2025benerf} such as E$^2$NeRF~\cite{qi2023e2nerf} and Ev-DeblurNeRF~\cite{cannici2024mitigating} combine blurry image formation~\cite{ma2022deblur} to model the camera motion during the exposure time while Ev-DeblurNeRF~\cite{cannici2024mitigating} further incorporates the EDI prior to provide additional color constraint to 3D NeRF. However, this color constraint often yields over-smoothed details by treating each RGB channel as brightness, which deviates from the real-world setting. In this work, we effectively restore intricate details by introducing a learnable brightness estimation function to the EDI formulation that better adapts to real-world settings.

\vspace{2mm}
\noindent \textbf{Diffusion-based image restoration.} Diffusion models have been successfully repurposed for image restoration tasks such as super-resolution (SR)~\cite{ho2022cascaded, li2022srdiff, saharia2022image, xia2023diffir, wang2024exploiting, yue2024resshift}, deblurring~\cite{whang2022deblurring, ren2023multiscale, chen2024hierarchical} and JPEG restoration~\cite{saharia2022palette, kawar2022jpeg}. A comprehensive summary of diffusion-based restoration methods can be found in this recent survey~\cite{li2023diffusion}.
DiffIR~\cite{xia2023diffir} is a two-stage training approach that generates a prior representation from a diffusion model to restore an image. DvSR~\cite{whang2022deblurring} surpasses regression-based methods by leveraging conditional diffusion models. More recently, diffusion prior has begun to be used for restoration of 3D representations such as NeRF. DiSR-NeRF~\cite{lee2024disr} first attempts to leverage a pretrained diffusion model to construct super-resolution (SR) NeRF by proposing a renoised variant of score distillation sampling~\cite{poole2022dreamfusion}, referred to as RSD optimization. In our approach, we propose a simpler framework to adopt RSD optimization compared to~\cite{lee2024disr}, further enhancing the edge details of the rendered image.

\begin{figure*}[tb]
\centering
  \includegraphics[width=1.0\linewidth]{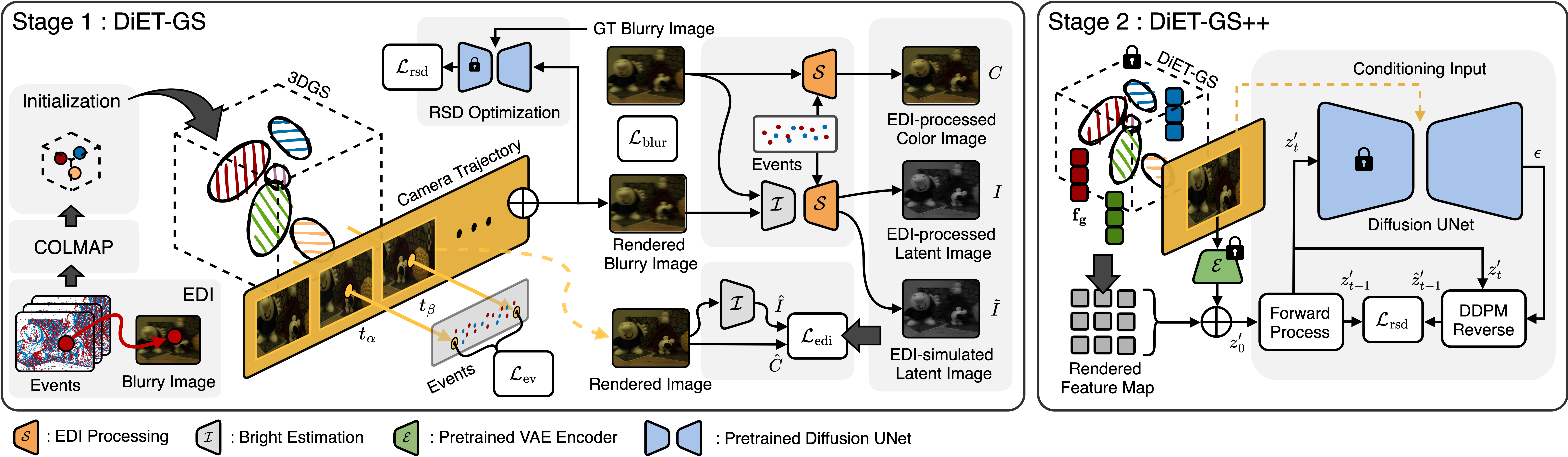}
  \vspace{-15pt}
  \caption{\textbf{Overall framework of our \ours.} \textcolor{emphasize}{\textbf{Stage 1 (\ours)}} optimizes the deblurring 3DGS with the event streams and diffusion prior. To preserve accurate color and clean details, we exploit the EDI prior in multiple ways, including color supervision $C$, guidance for fine-grained details $I$, and additional regularization $\tilde{I}$ with EDI simulation. \textcolor{emphasize}{\textbf{Stage 2 (\ourspp)}} is then employed to maximize the effect of diffusion prior by introducing extra learnable parameters $\mathbf{f}_{\mathbf{g}}$. \ourspp further refines the rendered images from \ours, effectively enhancing rich edge features. More details are explained in Sec.~\ref{sec:stage1} and Sec.~\ref{sec:stage2}.}
  \label{fig:framework}
  \vspace{-10pt}
\end{figure*}

\section{Preliminaries}
\noindent \textbf{Event Camera Model in Motion Deblurring.} Event is triggered when the absolute difference of logarithmic brightness between time $t_j$ and $t_{j-1}$ exceeds $\Theta_{p_j}$ at the same pixel location, where predefined threshold $\Theta_{p_j} \in \mathbb{R}^{+}$ controls the sensitivity to brightness change. It follows that:
\begin{equation}
\label{eq:event}
\small
\log(I(\mathbf{u}, t_j)) - \log(I(\mathbf{u}, t_{j-1})) = p_j \cdot \Theta_{p_j},
\end{equation}
where $I(\mathbf{u}, t)$ denotes instantaneous intensity  at a pixel $\mathbf{u}$ on a given time $t$. Polarity $p_j \in \{-1, 1\}$ specifies that either an increase or decrease in brightness. By denoting $\log(I(\mathbf{u}, t))$ as $L(t)$, Eq.~\ref{eq:event} can be generalized to any time interval $[t, t + \Delta t]$ by accumulating event signals as follows:
\begin{equation}
\label{eq:esi}
\vspace{-2mm}
\small
L(t+\Delta t) - L(t) = \Theta \cdot \mathbf{E}(t) = \Theta \int_{t}^{t+\Delta t} p\delta(\tau)\, \mathrm{d}\tau,
\end{equation}
where $\delta(\tau)$ is impulse function with unit integral and polarity $p_j$ on threshold $\Theta$ is omitted for brevity. By applying the $\exp(\cdot)$ function to both sides of Eq.~\ref{eq:esi}, we can rewrite $ I(t+\Delta t)=I(t) \cdot \exp(\Theta \cdot \mathbf{E}(t))$ to give the relationship between two instantaneous brightness observed at different time steps. 

We note that a blurry image taken from a conventional frame-based camera can be represented as averaging a sequence of sharp images $I$ acquired during the fixed exposure time $\tau$ as follows:
\begin{equation}
\label{eq:blur}
\small
\vspace{-2mm}
\mathbf{I}^{\mathbf{B}}(\mathbf{u}, t) = \frac{1}{\tau} \int_{t - \tau / 2}^{t + \tau / 2} I(\mathbf{u}, h)\, \mathrm{dh},
\end{equation}
where $\mathbf{I}^{\mathrm{B}}$ is the blurry image captured during the time interval $\Delta T = [t - \tau/2, t + \tau/2]$. 
By combining Eq.~\ref{eq:esi} and Eq.~\ref{eq:blur}, the connection between the blurry image $\mathbf{I}^{\mathbf{B}}$ and the latent sharp image $I$ at the same timestep $t$ can be constructed as:
\begin{equation}
\label{eq:edi}
\small
\mathbf{I}^{\mathbf{B}}(\mathbf{u}, t) = \frac{I(\mathbf{u}, t)}{\tau} \int_{t - \tau / 2}^{t + \tau / 2} \mathrm{exp}(\Theta \mathbf{E}(h))\, \mathrm{dh}.
\end{equation}
This relationship between the sharp image, blurry image, and event stream is known as Event-based Double Integral (EDI)~\cite{pan2019bringing}. Finally, we can remove motion blur from $\mathbf{I}^{\mathbf{B}}$ by solving for $I(\mathbf{u}, t)$ in Eq.~\ref{eq:edi} under the guidance of the event streams. In this paper, we propose a novel framework to leverage the EDI prior to constrain 3D Gaussian Splatting in recovering sharp rendered images with better fine-grained details and colors.

\vspace{2mm}
\noindent \textbf{Diffusion Prior in 3D.} 
Score-based diffusion models~\cite{song2020score, wang2023score} are a popular type of generative model that learns a score function that represents the gradient of the log probability density with respect to the data. Several works on Text-to-3D generation~\cite{poole2022dreamfusion, chen2023fantasia3d, huang2023dreamtime, lin2023magic3d, liu2023zero, metzer2023latent, wang2023score, wang2024prolificdreamer, zhu2023hifa} that leverage the score function of a pretrained diffusion model as a diffusion prior have shown remarkable results. Notably, Score Distillation Sampling (SDS)~\cite{poole2022dreamfusion} uses a score function of a diffusion model as an optimization target, providing the gradient from the score function of a pretrained diffusion model to guide differentiable image parametrization $\mathbf{z} = g(\theta)$ without retraining the diffusion model from scratch. In 3D, $\theta$ can be any learnable 3D representation~\cite{mildenhall2021nerf, kerbl20233d} with volume rendering function $g(\cdot)$. Recently, a variant of SDS has been introduced for image restoration task. Specifically, \cite{lee2024disr} proposes a renoised variant of SDS called Renoised Score Distillation (RSD) to use diffusion prior for super-resolution in 3D neural representation field.

Given a data sample $\mathbf{z}_0$, the forward process obtains the noisy latent $\mathbf{z}_t$ by adding Gaussian noise sample $\epsilon \in \mathcal{N}(0, \mathbf{I})$ at timestep $t$: $\mathbf{z}_t  = \sqrt{\bar{\alpha}_t}\mathbf{z}_0 + \sqrt{1 - \bar{\alpha}_t}\epsilon,$
where $\bar{\alpha}_t$ is timestep-dependent noising coefficient. In the DDPM reverse process, a diffusion U-net is trained to predict the noise $\epsilon(\mathbf{z},y,t)$ to denoise $\mathbf{z}_t$ into $\mathbf{z}_{t-1}$ as follows: $\mathbf{z}_{t-1} = \frac{1}{\sqrt{\alpha_t}}(\mathbf{z}_{t} - \frac{1 - \alpha_t}{\sqrt{1-\bar{\alpha}_t}}\epsilon(\mathbf{z}_t, y, t)) + \sigma_t\epsilon$,
where $y$ is the conditioning input and $\sigma_t$ is the standard deviation of Gaussian noise samples. Given the predicted denoised latent $\mathbf{\hat{z}}_{t-1}$ from $\mathbf{z}_t$ and the current noised latent $\mathbf{z}_{t-1}$ at timestep $t-1$, the objective 
of RSD is formulated as $\mathcal{L}_{rsd} = ||\mathbf{z}_{t-1} - \mathbf{\hat{z}}_{t-1}||.$ In this paper, we propose a simpler framework to leverage RSD optimization compared to~\cite{lee2024disr}.

\section{Our Method}
Fig.~\ref{fig:framework} shows an illustration of our DiET-GS framework which consists of two stages. Stage 1 (\cf Sec.~\ref{sec:stage1}) constructs a deblurring 3DGS by leveraging an EDI constraint derived from real-captured data and the prior knowledge of a pretrained diffusion model. Stage 2 (\cf Sec.~\ref{sec:stage2}) further refines the resulting images from Stage 1 by solely relying on diffusion prior to further enhance the edge details.

\subsection{Stage 1: DiET-GS}
\label{sec:stage1}

The goal of Stage 1 is to construct a set of 3D Gaussians from 3D Gaussian Splatting (3DGS)~\cite{kerbl20233d} to render sharp images from blurry images and event streams. We name the 3D Gaussians trained in this stage as \ours.

\vspace{2mm}
\noindent \textbf{Initialization.} The construction of 3DGS requires point cloud initialization and camera calibrations with structure-from-motion (SfM)~\cite{schonberger2016structure}, which often fail with blurry images. We mitigate this issue by leveraging EDI from Eq.~\ref{eq:edi} to recover an initial set of sharp images suitable for SfM. 
Specifically, given an RGB blurry image $\mathbf{C}^{\mathbf{B}}$ and an event stream $\mathbf{E}$ captured during an exposure time $\tau$, we reconstruct a sharp latent image $I$ from the grayscale of $\mathbf{C}^{\mathbf{B}}$ with EDI. This sharp image $I$ can then be warped to any timestep within the exposure period $[t - \tau/2, t + \tau/2]$ following $I(t + \Delta t) = I(t)\cdot\mathrm{exp}(\Theta \cdot \mathbf{E}(t))$ as shown in Eq.~\ref{eq:esi}. We obtain a set of latent images for each training view by warping the recovered latent image $I$ to each of the $n$ timesteps uniformly sampled from the exposure time. The recovered sharp latent images are subsequently fed into SfM for the estimation of the camera poses and point cloud.

\vspace{2mm}
\noindent \textbf{Blur Reconstruction Loss.} Let us denote the estimated camera poses along the approximated camera trajectory of a blurry image $\mathbf{C}^{\mathbf{B}}_i$ as $\mathbf{P}_i = \{p_{ij}\}^{n-1}_{j=0}$. Given the camera poses $\mathbf{P}_i$, we simulate the blurry image formation by discretizing Eq.~\ref{eq:blur} into:
\begin{equation}
\vspace{-2mm}
\label{eq:blur_discrete}
\small
\mathbf{\hat{C}}^{\mathbf{B}}_i = \frac{1}{n}\sum^{n-1}_{j=0}g_{\theta}(p_{ij}),
\end{equation}
where $g_{\theta}(\cdot)$ is the 3DGS with rendering function and $\mathbf{\hat{C}}^{\mathbf{B}}$ is the estimated motion-blurred image. 
We can now use the real captured blurry images $\{\mathbf{C}^{\mathbf{B}}\}^{k-1}_{i=0}$ to supervise the simulated blurry images $\{\mathbf{\hat{C}}^{\mathbf{B}}\}^{k-1}_{i=0}$, where $k$ is the number of training views.
Following the original 3DGS, we thus formulate a blur reconstruction loss to minimize the photometric error $\mathcal{L}_{\mathrm{P}}$ as $\mathcal{L}_{\mathrm{blur}} = \mathcal{L}_{\mathrm{p}}(\mathbf{C}^{\mathbf{B}}, \mathbf{\hat{C}}^{\mathbf{B}}) = (1-\lambda_1)\mathcal{L}_1 + \lambda_1 \mathcal{L}_{\mathrm{D-SSIM}}$.

\vspace{2mm}
\noindent \textbf{Event Reconstruction Loss.} Since an event stream provides blur-free microsecond-level measurements, it can be exploited to further aid fine-grained deblurring. To this end, we formulate an event reconstruction loss by leveraging the relation between brightness and generated events in Eq.~\ref{eq:esi}. Specifically, we synthesize the left-hand side of Eq.~\ref{eq:esi} by defining the simulated log brightness $\hat{L}(t_i)$ at time $t_i$ as:
\begin{equation}
\label{eq:intensity}
\small
\vspace{-0.5mm}
\hat{L}(t_i) = \log(h(\operatorname{CRF}(\hat{C}_i)))
\end{equation}
where $\hat{C}_i=g_{\theta}(p_i)$ is the sharp image rendered from the 3DGS at camera pose $p_i$ corresponding to time $t_i$, and $h(\cdot)$ is a luma conversion function implemented by following the BT.601~\cite{bt2011studio} standard. Following~\cite{cannici2024mitigating}, a learnable tone mapping function $\operatorname{CRF}(\cdot)$ is adopted to handle possible variations between the RGB and events response function.

To simulate the brightness change in Eq.~\ref{eq:esi}, we randomly sample two timesteps $t_{\alpha}$ and $t_{\beta}=t_{\alpha} + \Delta t$ along the camera trajectory and approximate the camera poses corresponding to the sampled timesteps via spherical linear interpolation~\cite{shoemake1985animating} of the known camera poses $\{\mathbf{P}_i\}^{k-1}_{i=0}$. Given the approximated camera poses at $t_{\alpha}$ and $t_{\beta}$, we can finally synthesize the brightness change $\Delta \hat{L}(t_{\alpha}, t_{\beta}) = \hat{L}(t_{\beta}) - \hat{L}(t_{\alpha})$ between time $t_{\alpha}$ and $t_{\beta}$ by estimating the log brightness from Eq.~\ref{eq:intensity}. Considering that the right-hand side of Eq.~\ref{eq:esi} can serve as ground-truth supervision, we thus formulate the event construction loss $\mathcal{L}_{\mathrm{ev}}$ as follows: $\mathcal{L}_{\mathrm{ev}} = ||\Delta \hat{L}(t_{\alpha}, t_{\beta}) - \Delta L(t_{\alpha}, t_{\beta})||^2_2,$
where $L(t_{\alpha}, t_{\beta})$ is the brightness difference observed from the event camera.

\vspace{2mm}
\noindent \textbf{Event Double Integration (EDI) Loss.} Although $\mathcal{L}_{\mathrm{ev}}$ is capable of producing sharp details to a certain degree, it lacks supervision in areas where events are not triggered. Furthermore, it is not trivial to blindly recover color details from an event response since the only color supervision comes from $\mathcal{L}_{\mathrm{blur}}$~\cite{cannici2024mitigating}. To this end, we propose a novel optimization problem that leverages EDI prior to further constrain the 3DGS in terms of 1) fine-grained details, 2) precise color and 3) regularizing the optimization.

Since EDI is defined in the monochrome brightness domain, we first model the EDI based on pixel intensity values. Specifically, we further exploit the composite function of the learnable camera response function $\operatorname{CRF}(\cdot)$ followed by $h(\cdot)$ in Eq.~\ref{eq:intensity} to estimate the brightness of given color images. Given the ground truth blurry image $\mathbf{C}^{\mathbf{B}}$, the brightness of $\mathbf{C}^{\mathbf{B}}$, denoted as $\mathbf{I}^{\mathbf{B}}$, is obtained by $\mathbf{I}^{\mathbf{B}} = h(\operatorname{CRF}(\mathbf{C}^{\mathbf{B}}))$. We then recover the mid-exposure pose of image $I$ from $\mathbf{I}^{\mathbf{B}}$ using Eq.~\ref{eq:edi}. Based on the latent image $I$, a sharp latent image $I_i$ at a randomly sampled timestep $t_i$ can be recovered by warping $I$ to timestep $t_i$ as stated in the initialization step. Given $I_i$ as image-level supervision, we synthesize the brightness of color image $\hat{C}_i$ rendered at the same timestep. Finally, the EDI loss in the monochrome pixel domain is formulated with the photometric error as follows: $\mathcal{L}_{\mathrm{edi\_gray}} := \mathcal{L}_{\mathrm{p}}(I_i, \hat{I}_i) = \mathcal{L}_{\mathrm{p}}(I_i, h(\operatorname{CRF}(\hat{C}_i))),$
where $\hat{I}_i$ is the estimated brightness of rendered color $\hat{C}_i$. The learnable bright response function $\operatorname{CRF}(\cdot)$ allows $\mathcal{L}_{\mathrm{edi\_gray}}$ to effectively restore fine-grained details for the overall image by naturally filling the gap between the rendered color space and the brightness change captured by the event sensor.

\begin{figure}[tb]
  \centering
  \includegraphics[width=0.8\linewidth]{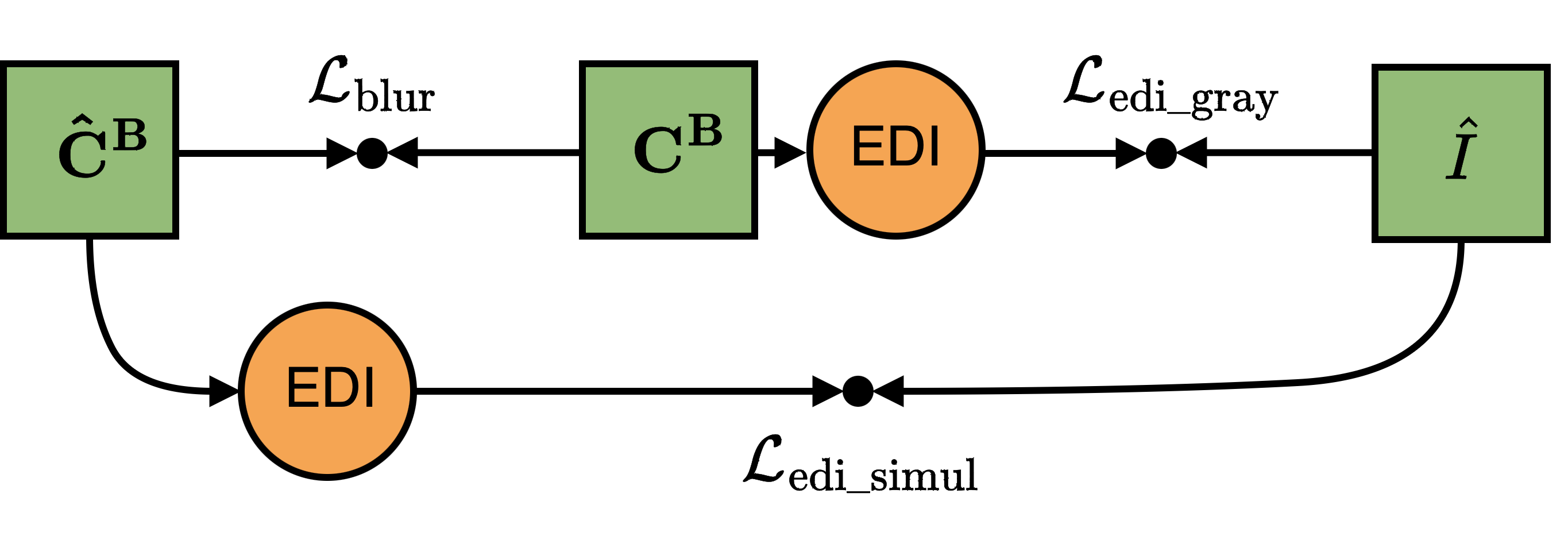}
  \vspace{-15pt}
  \caption{\textbf{Cycle consistency among the objective terms.} $\mathcal{L}_{\mathrm{edi\_simul}}$ follows the formulation of $\mathcal{L}_{\mathrm{edi\_gray}}$ except for substituting $\mathbf{C}^{\mathbf{B}}$ to simulated blurry image $\hat{\mathbf{C}}^{\mathbf{B}}$ derived from $\mathcal{L}_{\mathrm{blur}}$. It completes the cycle among the objective terms, further regularizing the fine-grained deblurring as shown in Fig.~\ref{fig:edi_simul_ablation}.}
  \label{fig:cycle}
\vspace{-15pt}
\end{figure}

Although effective in restoring fine-grained details, we empirically find that $\operatorname{CRF}(\cdot)$ often distorts color due to the lack of sufficient color supervision as shown in Fig.~\ref{fig:edi_ablation}. Inspired by~\cite{cannici2024mitigating}, we produce sharp RGB color $C$ by treating each RGB channel of $\mathbf{C}^{\mathbf{B}}$ as blurry brightness $\mathbf{I}^{\mathbf{B}}$ in Eq.~\ref{eq:edi} and applying channel-wise deblurring with EDI. Finally, sharp color $C_i$ at a randomly sampled timestep $t_i$ is obtained by warping $C$ to the corresponding timestep. Given $C_i$ as color supervision, the EDI loss in the RGB space can be formulated as $\mathcal{L}_{\mathrm{edi\_color}} = \mathcal{L}_{\mathrm{p}}(C_i, \hat{C}_i)$, where $\hat{C}_i = g_{\theta}(p_i)$ is the color image rendered at the same timestep as $C_i$.

In addition to the EDI loss described above for image-level supervision, we further leverage EDI for additional regularization.
Specifically, we synthesize both sides of Eq.~\ref{eq:edi} by simply replacing $\mathbf{I}^{\mathbf{B}}$ with the simulated blurry brightness $\mathbf{\hat{I}}^{\mathbf{B}} = h(\operatorname{CRF}(\mathbf{\hat{C}}^{\mathbf{B}}))$, which is the estimated brightness of $\mathbf{\hat{C}}^{\mathbf{B}}$. The EDI simulation loss can then be formulated as $\mathcal{L}_{\mathrm{edi\_simul}} = \mathcal{L}_{\mathrm{p}}(\tilde{I_i}, \hat{I_i})$, where $\tilde{I_i}$ is the sharp supervision obtained from the simulated blurry brightness via EDI processing. As illustrated in Fig.~\ref{fig:cycle}, $\mathcal{L}_{\mathrm{edi\_simul}}$ ensures cycle consistency among the objective terms, further facilitating the optimization as a regularization term.

Finally, the EDI loss is given by combining all of the EDI-based objectives as: $\mathcal{L}_{\mathrm{edi}} = \mathcal{L}_{\mathrm{edi\_gray}} + \mathcal{L}_{\mathrm{edi\_color}} + \mathcal{L}_{\mathrm{edi\_simul}}$.

\begin{figure*}[t]
\captionsetup[subfigure]{}
  \centering
  \begin{subfigure}{0.138\linewidth}\includegraphics[width=1.0\linewidth]{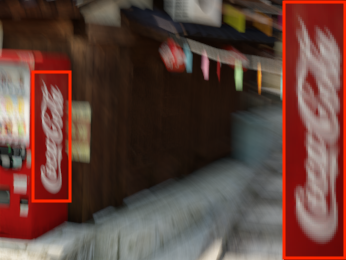}
  \end{subfigure}
  \begin{subfigure}{0.138\linewidth}\includegraphics[width=1.0\linewidth]{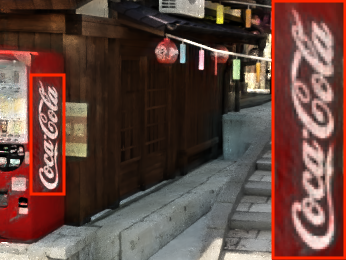}
  \end{subfigure}
  \begin{subfigure}{0.138\linewidth}\includegraphics[width=1.0\linewidth]{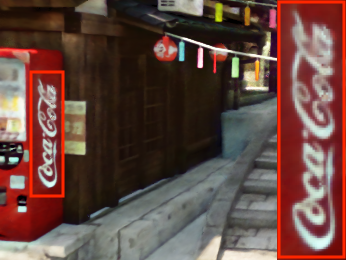}
  \end{subfigure}
  \begin{subfigure}{0.138\linewidth}\includegraphics[width=1.0\linewidth]{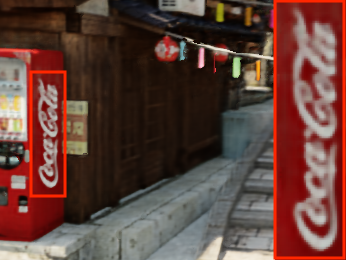}
  \end{subfigure}
  \begin{subfigure}{0.138\linewidth}\includegraphics[width=1.0\linewidth]{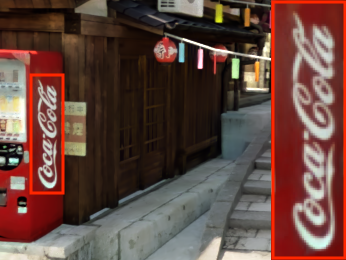}
  \end{subfigure}
  \begin{subfigure}{0.138\linewidth}\includegraphics[width=1.0\linewidth]{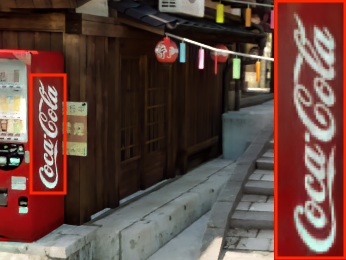}
  \end{subfigure}
  \begin{subfigure}{0.138\linewidth}\includegraphics[width=1.0\linewidth]{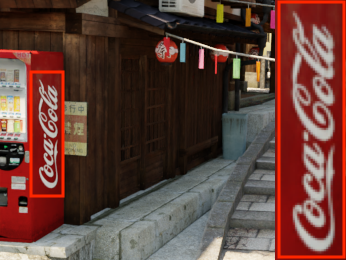}
  \end{subfigure}

  \centering
   \begin{subfigure}{0.138\linewidth}\includegraphics[width=1.0\linewidth]{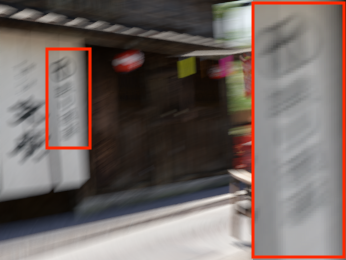}
  \end{subfigure}
  \begin{subfigure}{0.138\linewidth}\includegraphics[width=1.0\linewidth]{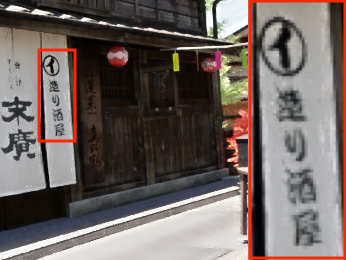}
  \end{subfigure}
  \begin{subfigure}{0.138\linewidth}\includegraphics[width=1.0\linewidth]{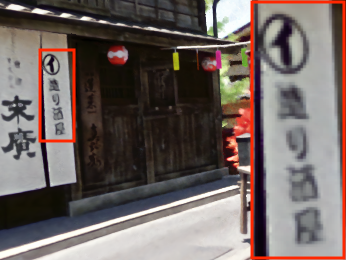}
  \end{subfigure}
  \begin{subfigure}{0.138\linewidth}\includegraphics[width=1.0\linewidth]{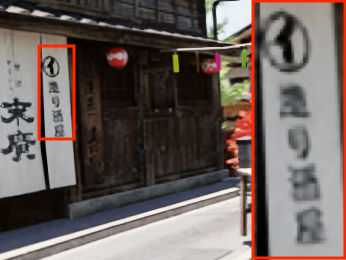}
  \end{subfigure}
  \begin{subfigure}{0.138\linewidth}\includegraphics[width=1.0\linewidth]{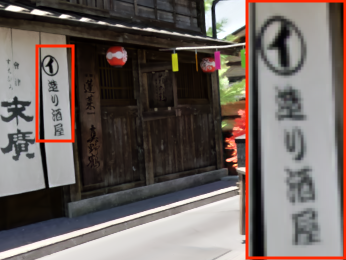}
  \end{subfigure}
  \begin{subfigure}{0.138\linewidth}\includegraphics[width=1.0\linewidth]{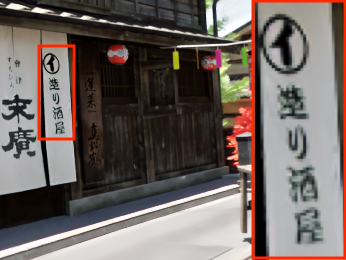}
  \end{subfigure}
  \begin{subfigure}{0.138\linewidth}\includegraphics[width=1.0\linewidth]{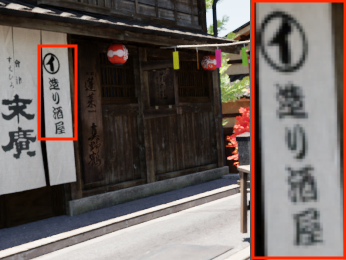}
  \end{subfigure}

  \begin{subfigure}{0.138\linewidth}\includegraphics[width=1.0\linewidth]{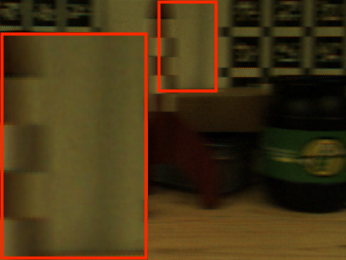}
  \end{subfigure}
  \begin{subfigure}{0.138\linewidth}\includegraphics[width=1.0\linewidth]{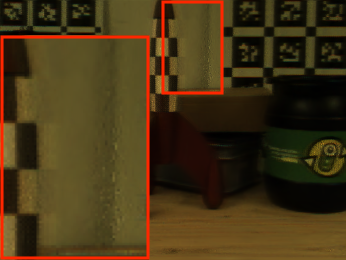}
  \end{subfigure}
  \begin{subfigure}{0.138\linewidth}\includegraphics[width=1.0\linewidth]{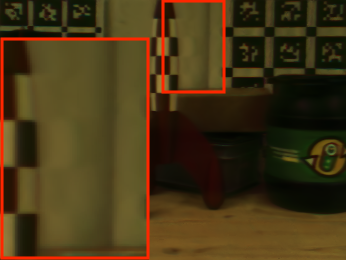}
  \end{subfigure}
  \begin{subfigure}{0.138\linewidth}\includegraphics[width=1.0\linewidth]{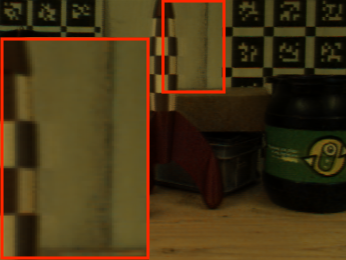}
  \end{subfigure}
  \begin{subfigure}{0.138\linewidth}\includegraphics[width=1.0\linewidth]{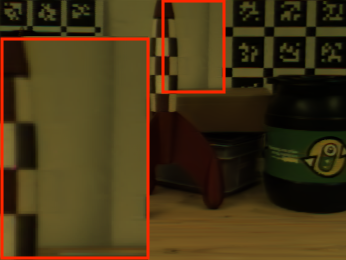}
  \end{subfigure}
  \begin{subfigure}{0.138\linewidth}\includegraphics[width=1.0\linewidth]{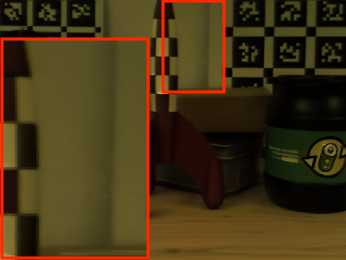}
  \end{subfigure}
  \begin{subfigure}{0.138\linewidth}\includegraphics[width=1.0\linewidth]{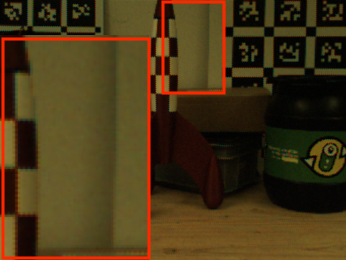}
  \end{subfigure}

  \begin{subfigure}{0.138\linewidth}\includegraphics[width=1.0\linewidth]{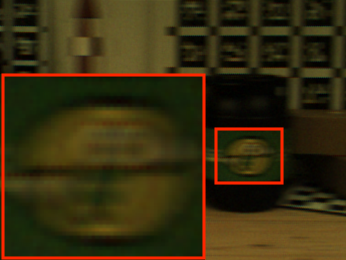}
  \subcaption{\scriptsize{Blur Image}}
  \end{subfigure}
  \begin{subfigure}{0.138\linewidth}\includegraphics[width=1.0\linewidth]{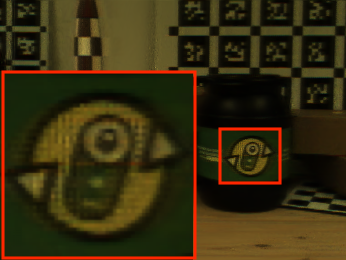}
  \subcaption{\scriptsize{EDI+GS~\cite{pan2019bringing}}}
  \label{fig:qualitative_b}
  \end{subfigure}
  \begin{subfigure}{0.138\linewidth}\includegraphics[width=1.0\linewidth]{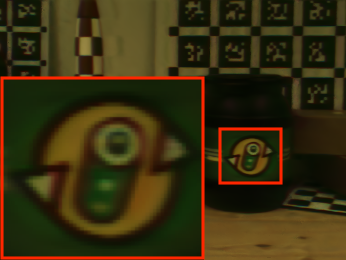}
  \subcaption{\scriptsize{E2NeRF~\cite{qi2023e2nerf}}}
  \label{fig:qualitative_c}
  \end{subfigure}
  \begin{subfigure}{0.138\linewidth}\includegraphics[width=1.0\linewidth]{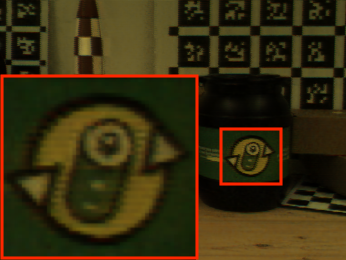}
  \subcaption{\scriptsize{Ev-DeblurNeRF~\cite{cannici2024mitigating}}}
  \label{fig:qualitative_d}
  \end{subfigure}
  \begin{subfigure}{0.138\linewidth}\includegraphics[width=1.0\linewidth]{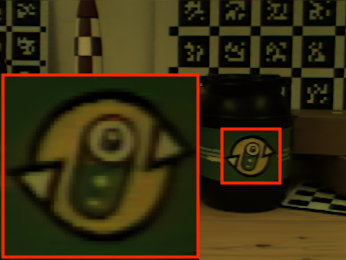}
  \subcaption{\scriptsize{\ours (Ours)}}
  \end{subfigure}
  \begin{subfigure}{0.138\linewidth}\includegraphics[width=1.0\linewidth]{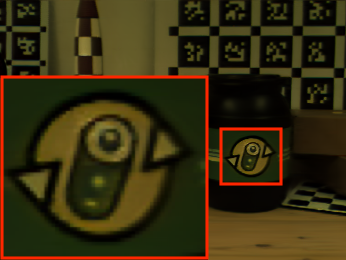}
  \subcaption{\scriptsize{\ourspp (Ours)}}
  \end{subfigure}
  \begin{subfigure}{0.138\linewidth}\includegraphics[width=1.0\linewidth]{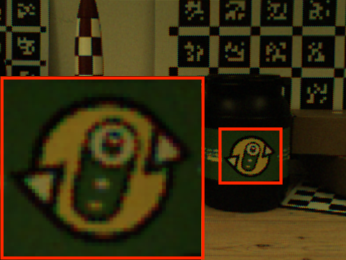}
  \subcaption{\scriptsize{GT}}
  \end{subfigure}
  \vspace{-5pt}
  \caption{\textbf{Qualitative comparisons on both synthetic (1st-2nd rows) and real-world (3rd-4th rows) datasets.} \ours shows cleaner texture with more accurate details compared to the event-based baselines while \ourspp further enhances these features with sharper definition, achieving the best visual quality.}
  \label{fig:qualitative}
  \vspace{-10pt}
\end{figure*}

\vspace{2mm}
\noindent \textbf{Leveraging Diffusion Prior.} Although event
streams can provide blur-free details, they are susceptible to
unnatural artifacts (Fig.~\ref{fig:qualitative_b}-\ref{fig:qualitative_d}) due to the unknown threshold $\Theta$ in Eq.~\ref{eq:esi}-\ref{eq:edi} and noise accumulated from the event~\cite{low2023robust}. Since pretrained diffusion model~\cite{rombach2022high} has already learned the distribution of natural images from large amounts of diverse datasets, it is intuitive to leverage this ``\textit{data-driven}'' prior in addition to our ``\textit{model-based}'' losses ($\mathcal{L}_{\mathrm{blur}}, \mathcal{L}_{\mathrm{ev}}$ and $ \mathcal{L}_{\mathrm{edi}}$) to further refine the output image more natural. Specifically, we adopt the RSD optimization strategy proposed in~\cite{lee2024disr} to our framework. However, unlike~\cite{lee2024disr}, our setting lacks the clean images which are necessary to guide noise prediction of diffusion model as conditional input. A straightforward solution is to utilize the sharp latent image $I$ from EDI processing as a surrogate for the clean image. Unfortunately, we empirically find that the EDI-processed images contain a lot of artifacts that are detrimental to the noise inference of the UNet in the diffusion model. We circumvent this issue by using the ground truth blurry images as an alternative.

We first render a blurry image $\hat{\mathbf{C}}^{\mathbf{B}}$ from Eq.~\ref{eq:blur}, and encode it to a latent $\mathbf{z}_0 = \mathcal{E}(\hat{\mathbf{C}}^{\mathbf{B}})$ via a pretrained VAE encoder $\mathcal{E}$. Subsequently, we apply the forward process of the diffusion model by introducing noise at timesteps $t$ and $t-1$ based on a predetermined noising schedule to get two noised latents $\mathbf{z}_t$ and $\mathbf{z}_{t-1}$. The UNet backbone~\cite{ronneberger2015u} of the pretrained diffusion model takes $\mathbf{z}_t$ as input and the ground truth blurry image $\mathbf{C}^{\mathbf{B}}$ as a condition to predict the noise residual $\mathbf{\epsilon}(\mathbf{z}_t, y, t)$. Given the predicted noise and $\mathbf{z}_{t}$, we then obtain the predicted denoised latent $\hat{\mathbf{z}}_{t-1}$ via the DDPM reverse process. Finally, the RSD loss $\mathcal{L}_{\mathrm{rsd}}$ is formulated as an L1 error between $\mathbf{z}_{t-1}$ and $\mathbf{\hat{z}}_{t-1}$. Since the input of the diffusion model $\hat{\mathbf{C}}^{\mathbf{B}}$ is obtained by averaging a set of rendered sharp images $\{\hat{C}\}^{n-1}_{i=0}$ along the camera trajectory, the supervision from the RSD loss is transferred to each rendered image, making them more natural.

\vspace{2mm}
\noindent \textbf{Training objective.} The final objective $\mathcal{L}_{s1}$ in Stage 1 is formulated as:
\begin{equation}
\label{eq:stage1_loss}
\small
\mathcal{L}_{\mathrm{s1}} = \lambda_{\mathrm{blur}} \mathcal{L}_{\mathrm{blur}} + \lambda_{\mathrm{ev}} \mathcal{L}_{\mathrm{ev}} + \lambda_{\mathrm{edi}} \mathcal{L}_{\mathrm{edi}} + \lambda_{\mathrm{rsd}} \mathcal{L}_{\mathrm{rsd}},
\end{equation}
with the weight $\lambda$ for each objective term. 

\begin{table*}[!h]
    \centering
    \footnotesize
    \resizebox{\linewidth}{!}{
    \begin{tabular}{c|c|c c c| c c | c c | >{\columncolor[gray]{0.8}}c >{\columncolor[gray]{0.8}}c}
    \hline
    \hline    
        \multirow{2}{*}{Dataset} & \multirow{2}{*}{Metric} & MPRNet+GS & EDI+GS & EFNet+GS & BAD-NeRF & BAD-GS & E2NeRF & Ev-DeblurNeRF & \ours & \ourspp \\
         & &\cite{zamir2021multi}&\cite{pan2019bringing} &\cite{sun2022event} &\cite{wang2023bad} &\cite{zhao2025bad} & \cite{qi2023e2nerf} & \cite{cannici2024mitigating} & (Ours) & (Ours) \\
         \hline
         \multirow{5}{*}{\makecell{EvDeblur\\-blender}}& PSNR$\uparrow$ & 18.76 & 23.69 & 21.03 & 19.78 & 22.23 & 24.54 & 24.76 & \textbf{26.69} & \underline{26.23} \\
         & SSIM$\uparrow$ &0.5912 & 0.7694 & 0.6413 & 0.6381 & 0.7213 & 0.7993 & 0.8038 & \textbf{0.8607} & \underline{0.8478} \\
         & LPIPS$\downarrow$ & 0.3545 & 0.1375 & 0.3214 & 0.2490 & 0.2012 & 0.1624 & 0.1788 & \underline{0.1064} & \textbf{0.1052} \\
         & MUSIQ$\uparrow$ &24.12 & 55.13 & 35.13 & 23.63 & 32.43 & 47.31 & 42.38 & \underline{57.67} & \textbf{59.91} \\
         & CLIP-IQA$\uparrow$ & 0.2413 & 0.2751 & 0.2314 & 0.1888 & 0.1993 & 0.2129 & 0.2300 & \underline{0.2769} & \textbf{0.2960} \\
         \hline
         \multirow{5}{*}{\makecell{EvDeblur\\-CDAVIS}}& PSNR$\uparrow$ & 27.51 & 32.95 & 30.97 & 28.47 & 29.12 & 31.54 & 32.30 & \textbf{34.22} & \underline{33.16} \\
         & SSIM$\uparrow$ & 0.7514 & 0.8922 & 0.8503 & 0.7981 & 0.8129 & 0.8687 & 0.8827 & \textbf{0.9223} & \underline{0.9039} \\
         & LPIPS$\downarrow$ & 0.2013 & 0.0790 & 0.1142 & 0.2526 & 0.2012 & 0.1059 & 0.0571 & \textbf{0.0496} & \underline{0.0502} \\
         & MUSIQ$\uparrow$ & 25.12 & 40.06 & 38.23 & 19.96 & 22.12 & 38.82 & 41.32 & \underline{45.80} & \textbf{50.44} \\
         & CLIP-IQA$\uparrow$ & 0.2134 & 0.2008 & 0.1934 & 0.1791 & 0.1812 & \underline{0.2235} & 0.2211 & 0.2072 & \textbf{0.2415} \\
         \bottomrule
    \end{tabular}}
    \newline
    \caption{\textbf{Quantitative comparisons on both synthetic and real-world dataset.} The results are the average of every scenes within the dataset. The best results are in \textbf{bold} while the second best results are \underline{underscored}.}
    \label{tab:quantitative}
    \vspace{-25pt}
\end{table*}

\subsection{Stage 2: DiET-GS++}
\label{sec:stage2}
Although the RSD optimization in Stage 1 allows \ours to produce more natural and precise color as shown in the 2nd row of Fig.~\ref{fig:edi_simul_ablation}, we empirically find that the performance improvement gained falls short of our expectation (\cf Tab.~\ref{tab:ablations}). We postulate that this is due to the joint optimization of the event-based loss which are $\mathcal{L}_{\mathrm{ev}}$ and $\mathcal{L}_{\mathrm{edi}}$, and the RSD loss. Event-based supervision is derived from real-captured event streams, which tend to optimize 3DGS for a specific training scene. In contrast, the RSD loss regularizes rendered images according to the distribution of diverse natural images modeled by a pretrained diffusion model. Jointly optimizing these two constraints reaches an equilibrium between scene-specific details guided by the event-based loss and the prior knowledge of the pretrained diffusion model. Consequently, it is likely to weaken the full effectiveness of $\mathcal{L}_{\mathrm{rsd}}$. To this end, we incorporate an additional training step to fully leverage the diffusion prior by introducing extra learnable parameters to \ours. We name the final model trained on Stage 2 as \ourspp.

Specifically, we attach a zero-initialized Gaussian feature $\mathbf{f}_{\mathbf{g}} \in \mathbb{R}^{D}$ for each 3D Gaussian $\mathbf{g}$ in the pretrained \ours, where $D$ is the feature dimension. Given a camera pose $p$, we use \ours to render a deblurred image $\hat{C}$ and a 2D feature map $\mathbf{f}_{\mathrm{2D}}$ by accumulating color and $\mathbf{f}_{\mathbf{g}}$, respectively. After encoding $\hat{C}$ to the latent $\mathbf{z}_0 = \mathcal{E}(\hat{C})$, we subsequently obtain a refined latent $\mathbf{z'}_0 = \mathbf{z}_0 + \mathbf{f}_{\mathrm{2D}}$ by combining $\mathbf{z}_0$ and $\mathbf{f}_{\mathrm{2D}}$. Gaussian noise samples at timesteps $t$ and $t-1$ are then introduced to $\mathbf{z'}_0$ to get noised latents $\mathbf{z'}_{t}$ and $\mathbf{z'}_{t-1}$. Finally, given $\hat{C}$ as conditional input, the UNet backbone of pretrained diffusion model predicts the noise residual of $\mathbf{z'}_{t}$ to derive the denoised latent $\hat{\mathbf{z}}'_{t-1}$. The RSD loss between $\mathbf{z'}_{t-1}$ and $\hat{\mathbf{z}}'_{t-1}$ is given as the optimization objective of Stage 2. We note that only the Gaussian features $\mathbf{f}_{\mathbf{g}}$ are trained during Stage 2, while the other parameters of \ours remain fixed to preserve the prior of event streams derived in Stage 1. After optimization, our model can render latent residual $\mathbf{f}_{\mathrm{2D}}$ which contains rich edge details directly guided by the pretrained diffusion model. In the inference, the final sharp image $\tilde{C}$ is obtained by decoding the refined latent $\mathbf{z}'_0$. Specifically, $\tilde{C} = \mathcal{D}(\mathbf{z}'_0) = \mathcal{D}(\mathbf{f}_{\mathrm{2D}} + \mathcal{E}(\hat{C}))$,
where $\mathcal{D}$ is pretrained VAE decoder (\cf Fig.~\ref{fig:teaser}). 

\vspace{2mm}
\noindent \textbf{Discussion.} Although Stage 2 of our framework uses RSD loss, we differ from~\cite{lee2024disr} as follows:
1) We exploit the rendering capability of 3DGS to obtain the learnable latent residual $\mathbf{f}_{\mathrm{2D}}$. In contrast, \cite{lee2024disr} manually creates $\mathbf{f}_{\mathrm{2D}}$ for all training poses and thus requires further synchronization of NeRF with the trained latent features after the RSD optimization. This synchronization stage is not necessary in our design since \ourspp can directly render the learned latent residual from the Gaussian features $\mathbf{f}_{\mathbf{g}}$, thus resulting in a simpler framework. Furthermore, our Stage 2 only takes $\leq20$ minutes of training time. 2) Our setting is more challenging than~\cite{lee2024disr} due to the lack of ground truth clean images to condition the diffusion UNet. In this regard, we utilize the image rendered from \ours as conditional input on the diffusion model to further enhance the edges.

\section{Experiments}

\subsection{Implementation Details}

We build our framework based on the official code of 3DGS~\cite{kerbl20233d} and DiSR-NeRF~\cite{lee2024disr}. Throughout Stage 1, we set the loss weights $\lambda_{\mathrm{blur}} = \lambda_{\mathrm{edi}} = \lambda_{\mathrm{rsd}} = 1.0$ and $\lambda_{\mathrm{ev}} = 0.1$, and execute 100,000 iterations of training. During Stage 2, training spans 2,000 iterations with a linearly decreasing time schedule of ancestral sampling. The number of poses $n$ estimated for each blurry image in the initialization of 3DGS is set to 9. All experiments are conducted using a single NVIDIA RTX 6000 GPU.

\subsection{Datasets.}
We evaluate our \ours and \ourspp on both synthetic and real-world datasets proposed by~\cite{cannici2024mitigating}. \textbf{EvDeblurBlender Dataset} is a synthetic dataset that consists of four synthetic scenes derived from DeblurNeRF~\cite{ma2022deblur}: \textit{factory}, \textit{pool}, \textit{tanabata} and \textit{trolley}. Blurry images are provided along with the corresponding synthetic event streams simulated by~\cite{rebecq2018esim}. Motion blur is produced during a 40ms exposure time with a single fast continuous motion by averaging a set of images rendered at 1000 FPS in linear RGB space. We set $\Theta = 0.2$ for the event threshold during training in synthetic scenes. The \textbf{EvDeblur-CDAVIS Dataset} contains five real-world scenes, each with 11 to 18 blurry training images paired with corresponding event streams. Color-DAVIS346~\cite{li2015design} is employed to capture both color events and standard frames at 346 $\times$ 260 pixel resolution using an RGBG Bayer pattern. A 1000ms exposure time is given to produce motion blur. The event threshold is set to $\Theta = 0.25$ for both negative and positive events during training in real-world scenes. Both synthetic and real-world datasets include five ground-truth (GT) sharp images captured from both seen and unseen viewpoints for each scene.

\subsection{Experiment Settings}
\noindent \textbf{Baseline.} Our baselines are divided into three categories. The first category is the naive combination of image deblurring methods and 3DGS, where images are initially deblurred and subsequently used as training views for 3DGS. Specifically, an image deblurring method MPRNet~\cite{zamir2021multi}, and event-based deblurring methods EDI~\cite{pan2019bringing} and EFNet~\cite{sun2022event} are adopted as baselines. The second category is the frame-only deblurring rendering methods, where only the RGB frames are utilized during training. We select BAD-NeRF~\cite{wang2023bad} and BAD-GS~\cite{zhao2025bad} for this category. The third category is event-based deblurring rendering methods, where the RGB frames and event streams are jointly leveraged. We select E2NeRF~\cite{qi2023e2nerf} and Ev-DeblurNeRF~\cite{cannici2024mitigating} as the most recent works in this category with publicly available code. To fairly compare with our methods, we utilize camera poses estimated from COLMAP for all baselines instead of using GT poses provided by the dataset.

\vspace{2mm}
\noindent \textbf{Evaluation Metrics.} We employ three standard metrics: Peak Signal-to-Noise Ratio (PSNR), Structural Similarity Index Measure (SSIM), and VGG-based Learned Perceptual Image Patch Similarity (LPIPS)~\cite{zhang2018unreasonable} to evaluate the quality of the predicted views on the target views. Since our framework produces rich edge details under the guidance of generative model, different predicted views can be valid for the same blurry image. We thus follow~\cite{yue2024resshift} to additionally adopt MUSIQ~\cite{ke2021musiq} and CLIP-IQA~\cite{wang2023exploring} as No-Reference Image Quality Assessment (NR-IQA) metrics to further evaluate the effectiveness of our framework.

\begin{table*}[!h]
    \centering
    \scriptsize
    \resizebox{\linewidth}{!}{
    \begin{tabular}{c c c c c c c | c c c c c c}
    \bottomrule
        $\mathcal{L}_{\mathrm{blur}}$ & $\mathcal{L}_{\mathrm{ev}}$ & $\mathcal{L}_{\mathrm{edi\_gray}}$ & $\mathcal{L}_{\mathrm{edi\_color}}$ & $\mathcal{L}_{\mathrm{edi\_simul}}$ & $\mathcal{L}_{\mathrm{rsd}}$ (S1) & $\mathcal{L}_{\mathrm{rsd}}$ (S2) & PSNR$\uparrow$ & SSIM$\uparrow$ & LPIPS$\downarrow$ & MUSIQ$\uparrow$  & CLIP-IQA$\uparrow$   \\
         \hline
         \ding{51} & &  &  &  &  & & 29.73 & 0.7797 & 0.2160 & 24.77  &  0.2031  \\
          \ding{51} & \ding{51} &  &  &  &  &  & 32.74 & 0.8460 & 0.1173 & 39.69  & \underline{0.2776}   \\
          
          \ding{51} & \ding{51} & \ding{51} &  &  &  &  & 33.91 & 0.8761 & 0.0752 & 44.25  & 0.2684   \\
          \ding{51} & \ding{51} & & \ding{51} &  &  &  & 34.48 & 0.8915 & 0.0826 & 40.79  & 0.2450   \\
          \ding{51} & \ding{51} & \ding{51} & \ding{51} &  &  &  & 34.92 & 0.9033 & 0.0624 & 43.79  & 0.2468   \\
          \ding{51} & \ding{51} & \ding{51} & \ding{51} & \ding{51} &  &  & \textbf{35.04} & \textbf{0.9068} & \textbf{0.0587} & 45.04  & 0.2490   \\
           \ding{51} & \ding{51} & \ding{51} & \ding{51} & 
           \ding{51} & \ding{51} &  & \underline{34.89} & \underline{0.9049} & \underline{0.0600} & \underline{45.37}  & 0.2584   \\
           \midrule
           \ding{51} & \ding{51} & \ding{51} & \ding{51} & \ding{51} & \ding{51} & \ding{51} & 33.86 & 0.8846 & 0.0634 & \textbf{51.71}  & \textbf{0.2955}   \\
           \bottomrule
    \end{tabular}}
    \newline
    \caption{\textbf{Ablation study on \ours and \ourspp}}
    \label{tab:ablations}
    \vspace{-25pt}
\end{table*}

\subsection{Quantitative Comparisons}
We show the quantitative results in Tab.~\ref{tab:quantitative}. Our \ours largely outperforms all baselines in PSNR, SSIM, and LPIPS on both synthetic and real-world datasets, showing the effectiveness of our framework to leverage EDI prior. Furthermore, our \ourspp shows significant improvement in MUSIQ and CLIP-IQA metrics, achieving the best results but showing a slight drop in PSNR and SSIM metrics. As already discussed in~\cite{yue2024resshift, lee2024disr}, since \ourspp is solely guided by a pretrained generative model, the resulting images may contain more variation with respect to GT samples compared to \ours which is supervised by real-captured data. Nonetheless, \ourspp still substantially improves the visual quality as shown in NR-IQA metrics. Qualitative comparisons in Sec.~\ref{sec:qualitative} further validate the effectiveness of our \ourspp. We also present the user study in the Supplementary material, thoroughly examining the efficacy of leveraging diffusion prior in Stage 2.

\begin{figure}[t]
\captionsetup[subfigure]{}
  \centering
  \begin{subfigure}{0.24\linewidth}\includegraphics[width=1.0\linewidth]{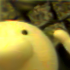}
  \end{subfigure}
  \begin{subfigure}{0.24\linewidth}\includegraphics[width=1.0\linewidth]{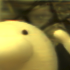}
  \end{subfigure}
  \begin{subfigure}{0.24\linewidth}\includegraphics[width=1.0\linewidth]{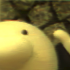}
  \end{subfigure}
  \begin{subfigure}{0.24\linewidth}\includegraphics[width=1.0\linewidth]{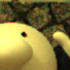}
  \end{subfigure}

  \begin{subfigure}{0.24\linewidth}\includegraphics[width=1.0\linewidth]{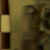}
  \subcaption{$\mathcal{L}_{\mathrm{edi\_gray}}$}
  \end{subfigure}
  \begin{subfigure}{0.24\linewidth}\includegraphics[width=1.0\linewidth]{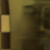}
  \subcaption{$\mathcal{L}_{\mathrm{edi\_color}}$}
  \label{fig:edi_ablation_b}
  \end{subfigure}
  \begin{subfigure}{0.24\linewidth}\includegraphics[width=1.0\linewidth]{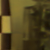}
  \subcaption{Both}
  \label{fig:edi_ablation_all}
  \end{subfigure}
  \begin{subfigure}{0.24\linewidth}\includegraphics[width=1.0\linewidth]{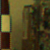}
  \subcaption{GT}
  \end{subfigure}
  \vspace{-5pt}
  \caption{\textbf{Ablation study on $\mathcal{L}_{\mathrm{edi\_gray}}$ and $\mathcal{L}_{\mathrm{edi\_color}}$}}
  \label{fig:edi_ablation}
  \vspace{-15pt}
\end{figure}

\subsection{Qualitative Comparisons}
\label{sec:qualitative}
Qualitative comparisons on both Ev-DeblurBlender and Ev-DeblurCDAVIS datasets are shown in Fig.~\ref{fig:qualitative}, making the following observations. 1) Relying solely on the EDI-preprocessed images to optimize the 3DGS yields unsatisfactory results, introducing numerous artifacts as shown in the 2nd column. While effective at recovering sharp edges, EDI often produces inaccurate details with relatively low visual quality, hindering effective optimization of the 3DGS. This highlights the need for additional objective terms proposed in our framework to further constrain the 3DGS. 2) We find that our \ours is capable of restoring cleaner textures and clearer edges compared to other baselines. For example, in the 3rd row, other baselines wrongly produce grid patterns on the wall stretched from the left object, while our \ours restores more accurate texture on the same part. Similarly, in the 4th row, color jittering from E2NeRF and artifacts from EDI+GS and Ev-DeblurNeRF are observed, while \ours exhibits cleaner details. The same observation is made for the text elements shown in the 1st and 2nd rows. 3) Our \ourspp further refines the textures and edge details generated from \ours, demonstrating the efficacy of fully leveraging diffusion prior in Stage 2. For instance, grid patterns on the wall in the 3rd row are further mitigated in \ourspp, while sharper edges are also observed in both text elements (1st-2nd rows) and non-text elements (4th row) compared to \ours. Overall, our final model \ourspp shows a marked improvement over the other baselines, consistently achieving more precise textures and well-defined details.

\subsection{Ablations}
In this section, we present various ablation studies to validate the contributions of each component proposed in our \ourspp. All the quantitative evaluations are conducted on a real-world scene, namely, \textit{Figures} sample.

\begin{figure}[t]
\captionsetup[subfigure]{}
  \centering
  \begin{subfigure}{0.32\linewidth}\includegraphics[width=1.0\linewidth]{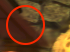}
  \subcaption{wo $\mathcal{L}_{\mathrm{edi\_simul}}$}
  \end{subfigure}
  \begin{subfigure}{0.32\linewidth}\includegraphics[width=1.0\linewidth]{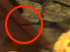}
  \subcaption{w $\mathcal{L}_{\mathrm{edi\_simul}}$}
  \end{subfigure}
  \begin{subfigure}{0.32\linewidth}\includegraphics[width=1.0\linewidth]{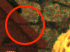}
  \subcaption{GT}
  \end{subfigure}
  \begin{subfigure}{0.32\linewidth}\includegraphics[width=1.0\linewidth]{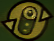}
  \subcaption{wo $\mathcal{L}_{\mathrm{rsd}}$ (S1)}
  \label{fig:rsd_ablation_a}
  \end{subfigure}
  \begin{subfigure}{0.32\linewidth}\includegraphics[width=1.0\linewidth]{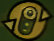}
  \subcaption{w $\mathcal{L}_{\mathrm{rsd}}$ (S1)}
  \label{fig:rsd_ablation_b}
  \end{subfigure}
  \begin{subfigure}{0.32\linewidth}\includegraphics[width=1.0\linewidth]{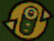}
  \subcaption{GT}
  \end{subfigure}
  
  \vspace{-5pt}
  \caption{\footnotesize{\textbf{Ablation on $\mathcal{L}_{\mathrm{edi\_simul}}$ (1st row) and $\mathcal{L}_{\mathrm{rsd}}$ (S1) (2nd row)}.}}
  \label{fig:edi_simul_ablation}
  \vspace{-15pt}
\end{figure}

\vspace{2mm}
\noindent \textbf{Event-based Supervision.} As shown in the 1st-2nd rows of Tab.~\ref{tab:ablations}, simulating brightness change in $\mathcal{L}_{\mathrm{ev}}$ improves PSNR by +3.01dB. Investigation of the EDI prior is conducted on the 3rd-6th rows of Tab.~\ref{tab:ablations} along with qualitative results on Fig.~\ref{fig:edi_ablation} and the 1st row of Fig.~\ref{fig:edi_simul_ablation}. As shown in the 1st row of Fig.~\ref{fig:edi_ablation}, using only $\mathcal{L}_{\mathrm{edi\_gray}}$ yields sharper details, \eg tiling on the background, and improves LPIPS scores compared to relying solely on $\mathcal{L}_{\mathrm{edi\_color}}$. However, color artifacts are also introduced due to the lack of sufficient color supervision. In contrast, $\mathcal{L}_{\mathrm{edi\_color}}$ generates accurate color with higher PSNR and SSIM scores, though some details are over-smoothed. Leveraging both $\mathcal{L}_{\mathrm{edi\_gray}}$ and $\mathcal{L}_{\mathrm{edi\_color}}$ yields the best performance on all the PSNR, SSIM, and LPIPS metrics among the 3rd to 5th rows of Tab.~\ref{tab:ablations}, achieving both precise color and clear details. The 2nd row of Fig.~\ref{fig:edi_ablation} further shows the synergistic effect between the two constraints, where more well-defined details are produced when they are jointly utilized. Similarly, adding EDI simulation $\mathcal{L}_{\mathrm{edi\_simul}}$ further aids fine-grained deblurring as shown in the red circles of Fig.~\ref{fig:edi_simul_ablation}, showing a +0.12dB increase in PSNR.

\vspace{2mm}
\noindent \textbf{Diffusion prior.} The following observations are made on the inclusion of diffusion prior. 1) The RSD optimization term in Stage 1 denoted as $\mathcal{L}_{\mathrm{rsd}}$ (S1) shows slight improvement in MUSIQ and CLIP-IQA by +0.33 and +0.0094, respectively. 2) $\mathcal{L}_{\mathrm{rsd}}$ (S1) can also serve as additional color guidance since the real-captured blurry image with ground-truth color information is given to the diffusion model as conditional input during RSD optimization, further constraining the color of \ours. As shown in the 2nd row of Fig.~\ref{fig:edi_simul_ablation}, our \ourspp guided by $\mathcal{L}_{\mathrm{rsd}}$ (S1) (\ref{fig:rsd_ablation_b}) generates better color compared to the version that omits RSD optimization in Stage 1 (\ref{fig:rsd_ablation_a}). 4) $\mathcal{L}_{\mathrm{rsd}}$ (S1) yields marginal performance improvement in NR-IQA metrics compared to the $\mathcal{L}_{\mathrm{rsd}}$ (S2). As discussed in Sec.~\ref{sec:stage2}, joint optimization of event-based objectives and RSD loss tends to weaken the full effect of diffusion prior. However, considering the first and second findings, we still decide to exploit diffusion prior in Stage 1 and further adopt Stage 2 with additional parameters to fully leverage the guidance from the pretrained diffusion model. 

\vspace{-5pt}
\section{Conclusion.}
\vspace{-5pt}

We present \ours, a novel framework to jointly use event streams and a pretrained score function. Our two-stage strategy allows 3DGS to recover clean and sharp images from motion blur. Our novel EDI constraints achieve both accurate color and fine-grained details, and the diffusion prior effectively enhances edge details with a simpler architecture. We believe our deblurring approach \ours have potentials for significant practical applications especially in low-light environment or fast-moving camera. 

\paragraph{Acknowledgement.}
This research / project is supported by the National Research Foundation (NRF) Singapore, under its NRF-Investigatorship Programme (Award ID. NRF-NRFI09-0008).

{
    \small
    \bibliographystyle{ieeenat_fullname}
    \bibliography{main}
}

\maketitlesupplementary
\appendix

In this supplementary material, we provide more implementation details, introduce additional evaluations on various tasks, and conduct further ablation studies with more qualitative and quantitative analysis.

\begin{itemize}
    \item More implementation details are provided in Sec.~\ref{sec:details}.
    \item We conduct the user study in Sec.~\ref{sec:user_study} to further evaluate the visual quality of our method.
    \item Additional evaluations on single image deblurring task is introduced in Sec.~\ref{sec:single_image_deblurring}.
    \item Further ablation studies for our \ourspp are presented in Sec.~\ref{sec:abl} to validate our design choice.
    \item More qualitative and quantitative results on novel-view synthesis are provided in Sec.~\ref{sec:qq}.
\end{itemize}

\section{Implementation Details}
\label{sec:details}

\noindent \textbf{Training.} During training, we follow the configuration of the original 3DGS. The learnable camera response function $\mathrm{CRF}(\cdot)$ is introduced after a 1,500-iteration warm-up. Similarly, the regularization term $\mathcal{L}_{\mathrm{edi\_simul}}$ is employed after a 7,000-iteration warm-up, since \ours should be able to simulate the blurry images properly. Following~\cite{cannici2024mitigating}, we leverage the color events in $\mathcal{L}_{\mathrm{ev}}$ during training on real-world scenes, where the color events record color intensity changes following a Bayer pattern~\cite{li2015design}. In this case, the luma conversion $h(\cdot)$ is dropped from Eq.~6 and $\mathcal{L}_{\mathrm{ev}}$ is directly applied to the color channel responsible for triggering events. Furthermore, since green pixels appear twice as often in an RGBG Bayer pattern, we weigh the events' contributions by 0.4, 0.2, and 0.4 for each of the RGB channels.

\vspace{2mm}
\noindent \textbf{Leveraging Diffusion Prior.} We use Stable Diffusion $\times$4 Upscaler (SD$\times$4)~\cite{rombach2022high} as a pretrained diffusion model to provide diffusion prior. SD$\times$4 is originally designed to upscale the image while recovering high-resolution details, with the low-resolution image as a conditional input to the diffusion UNet. However, we find that SD$\times$4 is also effective at enhancing edge details at the same resolution. During the RSD optimization, we sample uniform random crops of 128$\times$128 resolution in latent space for fast optimization speed, following~\cite{lee2024disr}. A constant learning rate of 1$e-$2 is employed for all learnable parameters $\mathbf{f}_{\mathbf{g}}$ in Stage 2.

\vspace{2mm}
\noindent \textbf{Color Correction.} As also noted in~\cite{choi2022perception, wang2024exploiting}, we empirically find that leveraging diffusion prior alone in Stage 2 can exhibit color shifts. To address this issue, we adopt wavelet-based color correction proposed in~\cite{wang2024exploiting} as a post-processing step. Specifically, let us denote the two images $\hat{C}$ and $\tilde{C}$ rendered from \ours and \ourspp respectively as follows:
\begin{equation}
\label{eq:render}
\small
\tilde{C} = \mathcal{D}(\mathbf{f}_{\mathrm{2D}} + \mathcal{E}(\hat{C})), \quad \hat{C} = g(p),
\end{equation}
where $g(\cdot)$ is the pretrained 3D Gaussians from \ours with a rendering function and $p$ is the given camera pose. We assume that $\hat{C}$ is capable of preserving the accurate color due to the color guidance from ground-truth blurry images and EDI prior in Stage 1. In contrast, $\tilde{C}$ from \ourspp tends to show a color shift since it solely relies on diffusion prior while the edge details are effectively enhanced. To combine the accurate color information from $\hat{C}$ and sharp edge details of $\tilde{C}$, we first decompose both images into high-frequency and low-frequency components via the wavelet decomposition. Considering that color information belongs to the low-frequency components while fine-grained details are mostly high-frequency components, we simply incorporate the low-frequency parts of $\hat{C}$ and high-frequency parts of $\tilde{C}$ to obtain the final output. More details about wavelet-based color correction can be found in~\cite{wang2024exploiting}.

\begin{figure}[t]
  \centering
  \begin{subfigure}{0.32\linewidth}\includegraphics[width=1.0\linewidth]{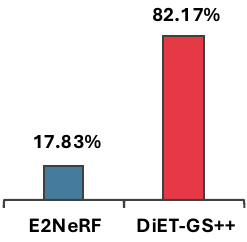}
  \label{fig:userstudy_a}
  \vspace{-10pt}
  \subcaption{}
  \end{subfigure}
  \begin{subfigure}{0.32\linewidth}\includegraphics[width=1.0\linewidth]{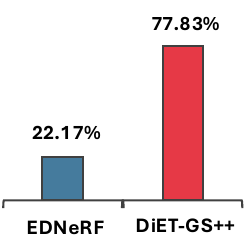}
  \label{fig:userstudy_b}
  \vspace{-10pt}
  \subcaption{}
  \end{subfigure}
  \begin{subfigure}{0.32\linewidth}\includegraphics[width=1.0\linewidth]{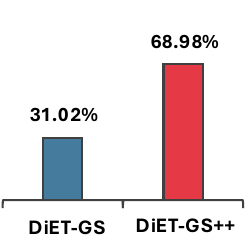}
  \vspace{-10pt}
  \subcaption{}
  \label{fig:userstudy_c}
  \end{subfigure}
  \vspace{-5pt}
  \caption{\textbf{User study.} \ourspp is compared to E2NeRF, Ev-DeblurNeRF (denoted as EDNeRF) and \ours by 60 evaluators for each pair. \ourspp gains significantly higher votes against the baselines, showing at least 37.96\% difference.}
  \label{fig:userstudy}
  \vspace{-10pt}
\end{figure}

\begin{table*}[t]
  \centering
  \small
  \resizebox{\linewidth}{!}{
  \begin{tabular}{l |c c | c c | c c | c c | c c | c c }
    \toprule
    \multirow{2}{*}{\textbf{Methods}} & \multicolumn{2}{c|}{\textit{Batteries}} & \multicolumn{2}{c|}{\textit{Powersupplies}} & \multicolumn{2}{c|}{\textit{Labequipment}} & \multicolumn{2}{c|}{\textit{Figures}} & \multicolumn{2}{c|}{\textit{Drones}} & \multicolumn{2}{c}{\textbf{\textit{Average}}} \\
     & MUSIQ$\uparrow$ & CLIP-IQA$\uparrow$ & MUSIQ$\uparrow$ & CLIP-IQA$\uparrow$ & MUSIQ$\uparrow$ & CLIP-IQA$\uparrow$ & MUSIQ$\uparrow$ & CLIP-IQA$\uparrow$ & MUSIQ$\uparrow$ & CLIP-IQA$\uparrow$ & MUSIQ$\uparrow$ & CLIP-IQA$\uparrow$ \\
    \midrule
    MPRNet~\cite{zamir2021multi} & 23.50 & 0.1074 & 34.01 & 0.1284 & 23.34 & 0.1082 & 24.10 & 0.1326 & 30.97 & 0.1034 & 27.18 & 0.1160 \\
    NAFNet~\cite{chen2022simple} & 34.43 & 0.2363 & 49.25 & 0.2045 & 35.47 & 0.2023 & 28.87 & 0.2264 & 41.64 & 0.1543 & 37.93 & 0.2047 \\
    Restormer~\cite{zamir2022restormer} & 28.71 & 0.1210 & 42.32 & 0.1154 & 32.13 & 0.1252 & 30.99 & 0.1631 & 36.37 & 0.0790 & 34.10 & 0.1207 \\
    \hline
    EDI~\cite{pan2019bringing} & 38.80 & 0.2013 & 49.63 & 0.2260 & 33.53 & 0.1338 & 41.77 & 0.2675 & 41.75 & 0.1991 & 41.10 & 0.2055 \\
    EFNet~\cite{sun2022event} & 35.32 & 0.1503 & 45.23 & 0.1934 & 30.13 & 0.1435 & 38.45 & 0.2234 & 39.12 & 0.1762 & 37.65 & 0.1773 \\
    BeNeRF~\cite{li2025benerf} & 47.31 & 0.1704 & 56.06 & 0.2445 & 44.25 & 0.1789 & 46.97 & 0.2653 & 48.71 & 0.2054 & 48.66 & 0.2129 \\
    \hline
    \rowcolor[gray]{0.8} \ourspp & \textbf{51.23} & \textbf{0.2654} & \textbf{56.32} & \textbf{0.2598} & \textbf{45.14} & \textbf{0.2034} & \textbf{52.43} & \textbf{0.3012} & \textbf{50.34} & \textbf{0.2078} & \textbf{51.09} & \textbf{0.2475} \\
    \bottomrule
  \end{tabular}
  }
  \caption{\textbf{Quantitative comparisons on single image deblurring with real-world datasets.}}
  \label{tab:single_deblurring}
  \vspace{-20pt}
\end{table*}

\begin{table}[t]
  \centering
  \scriptsize
  \resizebox{\linewidth}{!}{
  \begin{tabular}{l | c | c c c | c}
    \toprule
    \multirow{2}{*}{\textbf{Methods}} & \multirow{2}{*}{MUSIQ$\uparrow$} & \multicolumn{3}{c|}{Training time (hr)} & \multirow{2}{*}{Rendering time (s)} \\
     &  & Stage 1 & Stage 2 & Total &  \\
    \midrule
    E2NeRF~\cite{qi2023e2nerf} & 39.47 & 24.3 & - & 24.3 & 2.4139   \\
    Ev-DeblurNeRF~\cite{cannici2024mitigating} & 39.70 & 3.4 & - & 3.4 & 0.8861\\
    \midrule
    \rowcolor[gray]{0.8} \ours & 45.31 & 9.8 & - & 9.8 & \textbf{0.0014} \\
    \rowcolor[gray]{0.8} \ourspp & \textbf{51.71} & 9.8 & 0.17 & 10.0 & 1.8703 \\
    \rowcolor[gray]{0.8} \ourspp-\textit{light} & 50.23 & 1.1 & 0.17 & \textbf{1.3} & 1.8703 \\
    \bottomrule
  \end{tabular}}
  \caption{\textbf{Comparison on training time and rendering time.}}
   \label{tab:time}
   \vspace{-20pt}
\end{table}

\section{User Study}
\label{sec:user_study}

To evaluate the visual quality in terms of human perception, we conduct a user study with 60 evaluators. Specifically, we collect 30 pairs of samples from the test views of both synthetic and real-world datasets, where each pair consists of two images rendered from identical poses using different methods. During the user study, evaluators were asked to select the image with better quality between the two presented options for every pair.

\vspace{2mm}
\noindent \textbf{Baselines.} We compare our \ourspp to event-based deblurring rendering methods, including E2NeRF~\cite{qi2023e2nerf} and Ev-DeblurNeRF~\cite{cannici2024mitigating}. Furthermore, \ourspp is also compared with \ours trained from Stage 1 to demonstrate the efficacy of leveraging diffusion prior in Stage 2.

\vspace{2mm}
\noindent \textbf{Results.} As shown in Fig.~\ref{fig:userstudy}, our \ourspp gains at least 68.98\% of the votes in each comparison, further validating the effectiveness of our framework. It also shows the clear gap of 37.96\% over \ours (\cf Fig.~\ref{fig:userstudy_c}), highlighting the efficacy of enhancing the edge details with diffusion prior in Stage 2.

\section{Single Image Deblurring}
\label{sec:single_image_deblurring}

We also conduct experiments on the single image deblurring task using the real-world Ev-DeblurCDAVIS dataset~\cite{cannici2024mitigating}. For evaluation, we randomly select 5 blurry images per scene and compare our \ourspp against various single image deblurring baselines on these sampled images.

\vspace{2mm}
\noindent \textbf{Baselines.} We classify the baselines into three categories. The first category is frame-based single image deblurring methods that rely solely on RGB frames to recover a clean image. MPRNet~\cite{zamir2021multi}, NAFNet~\cite{chen2022simple}, and Restormer~\cite{zamir2022restormer} are selected for this category. The second category is event-enhanced deblurring methods that utilize additional event data to improve visual quality, consisting of EDI~\cite{pan2019bringing} and EFNet~\cite{sun2022event}. The third category combines NeRF and events to tackle single image deblurring, where BeNeRF~\cite{li2025benerf} is chosen for this category. BeNeRF reconstructs the 3D scenes by learning the camera trajectory from a single blurry image and corresponding event stream to deblur the given single view. Once we have trained BeNeRF, the deblurred image is produced by rendering the mid-exposure pose of the image along the estimated camera trajectory.

\vspace{2mm}
\noindent \textbf{Evaluation metrics.} Since real-world dataset lacks the ground-truth images for the mid-exposure poses of blurry views, we employ the No Reference Image Quality Assessment (NR-IQA) metrics: MUSIQ~\cite{ke2021musiq} and CLIP-IQA~\cite{wang2023exploring} for the evaluation.

\vspace{2mm}
\noindent \textbf{Results.} We present the quantitative comparisons in Tab.~\ref{tab:single_deblurring}. Our DiET-GS++ consistently outperforms all baselines in every 5 real-world scenes. Specifically, compared to BeNeRF, performance is improved by an average of 2.43 and 0.0346 in MUSIQ and CLIP-IQA scores, respectively. Furthermore, we also present qualitative comparisons in Fig.~\ref{fig:quaitative_single}. As shown in 2nd column, frame-based image deblurring method NAFNet often produces inaccurate details since it solely relies on blurry images to recover fine-grained details. EDI and BeNeRF recover more precise details, benefiting from the event-based cameras while severe artifacts are still exhibited. Our \ourspp shows the best visual quality with cleaner and well-defined details by leveraging EDI and pretrained diffusion model as prior.

\section{Ablation Study}
\label{sec:abl}

We present additional ablation studies to thoroughly investigate each component of \ourspp. All the experiments are conducted on a real-world scene, namely, \textit{Figures} sample.

\subsection{Training and Rendering Efficiency.}
We compare the optimization and rendering efficiency of our method to event-enhanced rendering methods, including E2NeRF~\cite{qi2023e2nerf} and Ev-DeblurNeRF~\cite{cannici2024mitigating} in Tab.~\ref{tab:time}. We present the training time of Stage 1 and Stage 2 separately, while the training time of Stage 2 remains blank if the corresponding method employs single-stage training. We observe from Tab.~\ref{tab:time}: 1) \ours and \ourspp require longer training time compared to Ev-DeblurNeRF. We find that RSD optimization in Stage 1 is the main factor of prolonged training time, since the gradient from the RSD loss flows to the 3D Gaussians through the pretrained VAE encoder, which introduces significant computational overhead. We thus propose the light variant of our \ourspp in the 5th row by simply excluding the RSD loss in Stage 1, which we refer to as \ourspp-\textit{light}. Despite a slight performance drop in MUSIQ scores, our variant \ourspp-\textit{light} shows the fastest optimization speed with a $\times$2.6 speedup in convergence compared to Ev-DeblurNeRF. 2) Training time for Stage 2 in \ourspp only requires 0.17 hours, while showing a significant improvement in MUSIQ scores compared to \ours. In contrast to RSD optimization in Stage 1, the learnable latent residual is directly rendered from \ours without exploiting the VAE encoder, which thus leads to faster gradient computation. 3) \ours enables real-time rendering, benefiting from the explicit representations of 3DGS. However, our \ourspp exhibits longer rendering time compared to Ev-DeblurNeRF, since the rendered image is further refined through the VAE encoder and decoder. Nonetheless, leveraging diffusion prior is reasonable given the performance improvement of 12.0 MUSIQ scores compared to Ev-DeblurNeRF.

\begin{table}[t]
  \centering
  \scriptsize
  \resizebox{\linewidth}{!}{
  \begin{tabular}{c | c c c | c c}
    \toprule
    $\mathrm{CRF(\cdot)}$ & PSNR$\uparrow$ & SSIM$\uparrow$ & LPIPS$\downarrow$ & MUSIQ$\uparrow$ & CLIP-IQA$\uparrow$ \\
    \midrule
    \ding{56} & 32.93 & 0.8703 & 0.1123 & 38.93 & 0.2000   \\
    \ding{51} & \textbf{34.89} & \textbf{0.9049} & \textbf{0.0600} & \textbf{45.31} & \textbf{0.2471}  \\
    \bottomrule
  \end{tabular}}
  \caption{\textbf{Ablation on camera response function $\mathrm{CRF(\cdot)}$.}}
   \label{tab:crf}
   \vspace{-15pt}
\end{table}

\begin{figure}[t]
\captionsetup[subfigure]{}
  \centering
  \begin{subfigure}{0.49\linewidth}\includegraphics[width=1.0\linewidth]{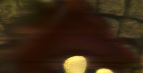}
  \subcaption{wo $\mathrm{CRF(\cdot)}$}
  \end{subfigure}
  \begin{subfigure}{0.49\linewidth}\includegraphics[width=1.0\linewidth]{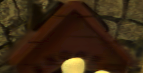}
  \subcaption{w $\mathrm{CRF(\cdot)}$}
  \end{subfigure}
  \vspace{-5pt}
  \caption{\textbf{Qualitative analysis on camera response function.}}
  \label{fig:crf}
  \vspace{-5pt}
\end{figure}

\begin{figure}[t]
\captionsetup[subfigure]{}
  \centering
  \begin{subfigure}{0.32\linewidth}\includegraphics[width=1.0\linewidth]{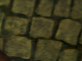}
  \subcaption{wo color correction}
  \end{subfigure}
  \begin{subfigure}{0.32\linewidth}\includegraphics[width=1.0\linewidth]{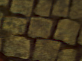}
  \subcaption{w color correction}
  \end{subfigure}
  \begin{subfigure}{0.32\linewidth}\includegraphics[width=1.0\linewidth]{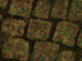}
  \subcaption{GT}
  \end{subfigure}
  \vspace{-5pt}
  \caption{\textbf{Ablation on wavelet-based color correction.}}
  \label{fig:cc}
  \vspace{-10pt}
\end{figure}

\subsection{Camera Response Function}
Tab.~\ref{tab:crf} shows the effectiveness of leveraging the learnable camera response function $\mathrm{CRF(\cdot)}$, showing the performance improvement in all 5 metrics. Furthermore, Fig.~\ref{fig:crf} demonstrates that the $\mathrm{CRF(\cdot)}$ function is capable of restoring more well-defined details. As noted in~\cite{cannici2024mitigating}, employing the learnable camera response function naturally fills the gap between the RGB space and the brightness change perceived by the event camera, thus effectively restoring the intricate details.

\vspace{2mm}
\noindent \textbf{Discussion.} Although we adopt the similar strategy with~\cite{cannici2024mitigating} by leveraging learnable camera response function, we differ from~\cite{cannici2024mitigating} as follows: We further combine the $\mathrm{CRF(\cdot)}$ function into the EDI formulation, proposing the novel EDI constraint for enhancing fine-grained details. As shown in the Fig. 5b, EDI color guidance $\mathcal{L}_{\mathrm{edi\_color}}$ proposed by~\cite{cannici2024mitigating} often yields over-smoothed details since it treats each RGB channel as brightness which deviates from the real-world setting. To compensate the well-defined details, we propose $\mathcal{L}_{\mathrm{edi\_gray}}$ by modeling the EDI in the brightness domain with exploiting learnable $\mathrm{CRF(\cdot)}$ function. Using the $\mathcal{L}_{\mathrm{edi\_color}}$ and $\mathcal{L}_{\mathrm{edi\_gray}}$ together enables the mutual compensation between accurate color and well-defined details, achieving the best visual quality as shown in the Fig.~\ref{fig:edi_ablation_all}.

\begin{table}[t]
  \centering
  \scriptsize
  \resizebox{\linewidth}{!}{
  \begin{tabular}{c | c c c | c c}
    \toprule
    Conditional input & PSNR$\uparrow$ & SSIM$\uparrow$ & LPIPS$\downarrow$ & MUSIQ$\uparrow$ & CLIP-IQA$\uparrow$ \\
    \midrule
    EDI-processed image & 34.25 & 0.8964 & 0.0754 & 41.47 & 0.2233   \\
    ground-truth blurry image & \textbf{34.89} & \textbf{0.9049} & \textbf{0.0600} & \textbf{45.31} & \textbf{0.2471}  \\
    \bottomrule
  \end{tabular}}
  \caption{\scriptsize{\textbf{Ablation on conditional input of RSD optiimzation in Stage 1.}}}
   \label{tab:cond}
   \vspace{-20pt}
\end{table}
  
\subsection{Conditional Input of Diffusion UNet}

In Tab.~\ref{tab:cond}, we explore the various options for conditional input of the diffusion UNet during the RSD optimization in Stage 1. The 1st row of Tab.~\ref{tab:cond} exploits the EDI-processed image as conditional input, while the sharp image rendered from 3DGS is given as the input to the diffusion process. However, this choice leads to inferior performance compared to leveraging the ground-truth blurry image as conditional input (2nd row). We postulate that this is because the unnatural artifacts introduced by EDI are often detrimental to the noise inference of the diffusion UNet. Despite the motion blur in the image, the ground-truth blurry image provides more natural guidance to noise prediction, such as accurate color prior, since it is real-captured from the frame-based camera.

\subsection{Wavelet-based Color Correction}

Fig.~\ref{fig:cc} presents the effectiveness of wavelet-based color correction. It effectively mitigates the color shift introduced from diffusion prior, achieving better color.

\section{More Results on Novel-View Synthesis}
\label{sec:qq}

\noindent \textbf{Quantitative Results.}
In Tab.~\ref{tab:quantitative_real} and Tab.~\ref{tab:quantitative_synthetic}, we present the additional quantitative results on novel-view synthesis for each scene in both real-world and synthetic datasets. In most cases, our \ours and \ourspp significantly outperform existing baselines across all five evaluation metrics, showing the effectiveness of our framework.

\vspace{2mm}
\noindent \textbf{Qualitative Results.}
In Fig.~\ref{fig:quaitative_real} and Fig.~\ref{fig:qualitative_synthetic}, we present more qualitative comparisons on novel-view synthesis in both real-world and synthetic datasets. Our \ourspp is capable of restoring: 1) \textit{accurate color}, 2) \textit{fine-grained details} and 3) \textit{clean texture}, thus achieving the best visual quality compared to the existing baselines.

\section{Limitation}

Following previous works~\cite{cannici2024mitigating, qi2023e2nerf}, we structure \ours assuming uniform-speed camera motion and dense, low-noise events. While real-world scenarios may not always meet these ideal conditions, advanced techniques like~\cite{low2023robust} could extend our method's applicability.

\begin{table*}[!h]
    \centering
    \footnotesize
    \resizebox{\linewidth}{!}{
    \begin{tabular}{c|c|c c c| c c | c c | >{\columncolor[gray]{0.8}}c >{\columncolor[gray]{0.8}}c}
    \hline
    \hline    
        \multirow{2}{*}{Scene} & \multirow{2}{*}{Metric} & MRPNet+GS & EDI+GS & EFNet+GS & BAD-NeRF & BAD-GS & E2NeRF & Ev-DeblurNeRF & \ours & \ourspp \\
         & &\cite{zamir2021multi}&\cite{pan2019bringing} &\cite{sun2022event} &\cite{wang2023bad} &\cite{zhao2025bad} & \cite{qi2023e2nerf} & \cite{cannici2024mitigating} & (Ours) & (Ours) \\
         \hline
         \multirow{5}{*}{\textit{batteries}}& PSNR$\uparrow$ & 28.42 & 33.11 & 31.30 & 28.29 & 28.73 & 31.49 & 32.63 & \textbf{34.52} & \underline{33.51} \\
         & SSIM$\uparrow$ & 0.7518 & 0.8994 & 0.8556 & 0.8086 & 0.8217 & 0.8715 & 0.8938 & \textbf{0.9304} & \underline{0.9118} \\
         & LPIPS$\downarrow$ & 0.1948 & 0.0613 & 0.0804 & 0.2245 & 0.1651 & 0.0932 & \underline{0.0443} & \textbf{0.0435} & 0.0444 \\
         & MUSIQ$\uparrow$ & 22.13 & 37.90 & 35.51 & 17.71 & 20.20 & 37.48 & 42.99 & \underline{45.66} & \textbf{49.89} \\
         & CLIP-IQA$\uparrow$ & 0.2338 & 0.2182 & 0.2293 & 0.1887 & 0.1918 & \underline{0.2445} & 0.2292 & 0.2327 & \textbf{0.2603} \\
         \hline
         \multirow{5}{*}{\textit{figures}}& PSNR$\uparrow$ & 28.18 & 33.51 & 31.28 & 29.31 & 30.12 & 32.59 & 32.82 & \textbf{34.89} & \underline{33.86} \\
         & SSIM$\uparrow$ & 0.7311 & 0.8723 & 0.8317 & 0.7703 & 0.7767 & 0.8543 & 0.8577 & \textbf{0.9049} & \underline{0.8846} \\
         & LPIPS$\downarrow$ & 0.2146 & 0.0977 & 0.1324 & 0.2935 & 0.2438  & 0.1108 & 0.0687 & \textbf{0.0600} & \underline{0.0634} \\
         & MUSIQ$\uparrow$ & 23.45 & 38.48 & 37.13 & 19.50 & 22.13 & 39.47 & 39.70 & \underline{45.37} & \textbf{51.71} \\
         & CLIP-IQA$\uparrow$ & 0.2418 & 0.2384 & 0.2218 & 0.1836 & 0.1898 & \underline{0.2624} & 0.2441 & 0.2584 & \textbf{0.2955} \\
         \hline    
         \multirow{5}{*}{\textit{drones}}& PSNR$\uparrow$ & 27.13 & \underline{33.02} & 31.18 & 28.51 & 29.19 & 31.03 & 31.62 & \textbf{34.08} & 32.92 \\
         & SSIM$\uparrow$ & 0.7634 & 0.9025 & 0.8617 & 0.8123 & 0.8317 & 0.8780 & 0.8866 & \textbf{0.9339} & \underline{0.9152} \\
         & LPIPS$\downarrow$ & 0.2012 & 0.0832 & 0.1293 & 0.2122 & 0.1687 & 0.1075 & 0.0538 & \textbf{0.0387} & \underline{0.0396} \\
         & MUSIQ$\uparrow$ & 28.38 & 42.35 & 41.18 & 19.05 & 22.20 & 39.00 & 41.81 & \underline{47.58} & \textbf{50.17} \\
         & CLIP-IQA$\uparrow$ & 0.1718 & 0.1633 & 0.1526 & 0.1723 & 0.1743 & \underline{0.1877} & 0.1773 & 0.1778 & \textbf{0.2028} \\
         \hline    
         \multirow{5}{*}{\textit{powersupplies}}& PSNR$\uparrow$ & 26.37 & 32.10 & 30.92 & 27.35 & 28.38 & 31.06 & 32.05 & \textbf{33.54} & \underline{32.37} \\
         & SSIM$\uparrow$ & 0.7513 & 0.8955 & 0.8516 & 0.7953 & 0.8071 & 0.8820 & 0.8980 & \textbf{0.9271} & \underline{0.9108} \\
         & LPIPS$\downarrow$ & 0.1824 & 0.0657 & 0.1029 & 0.2756 & 0.2247 & 0.0826 & 0.0492 & \underline{0.0460} & \textbf{0.0459} \\
         & MUSIQ$\uparrow$ & 31.48 & 46.04 & 44.15 & 24.68 & 24.90 & 45.17 & 47.97 & \underline{50.25} & \textbf{55.83} \\
         & CLIP-IQA$\uparrow$ & 0.2477 & 0.2307 & 0.2219 & 0.1762 & 0.1701 & 0.2373 & \underline{0.2501} & 0.2078 & \textbf{0.2531} \\
         \hline
         \multirow{5}{*}{\textit{labequipment}}& PSNR$\uparrow$ & 27.47 & 33.00 & 30.18 & 28.89 & 29.19 & 31.51 & 32.36 & \textbf{34.06} & \underline{33.13} \\
         & SSIM$\uparrow$ & 0.7598 & 0.8911 & 0.8512 & 0.8042 & 0.8276 & 0.8578 & 0.8772 & \textbf{0.9150} & \underline{0.8971} \\
         & LPIPS$\downarrow$ & 0.2138 & 0.0871 & 0.1262 & 0.2563 & 0.2037 & 0.1355 & 0.0696 & \underline{0.0599} & \textbf{0.0575} \\
         & MUSIQ$\uparrow$ & 20.18 & 35.54 & 33.18 & 18.84 & 21.19 & 32.95 & 34.14 & \underline{40.21} & \textbf{44.60} \\
         & CLIP-IQA$\uparrow$ & 0.1722 & 0.1534 & 0.1418 & 0.1749 & 0.1804 & 0.1854 & \textbf{0.2048} & 0.1708 & \underline{0.1958} \\
         \bottomrule
    \end{tabular}}
    \newline
    \caption{\textbf{Quantitative comparisons on novel-view synthesis in 5 real-world scenes}}
    \label{tab:quantitative_real}
\end{table*}

\begin{table*}[!h]
    \centering
    \footnotesize
    \resizebox{\linewidth}{!}{
    \begin{tabular}{c|c|c c c| c c | c c | >{\columncolor[gray]{0.8}}c >{\columncolor[gray]{0.8}}c}
    \hline
    \hline    
        \multirow{2}{*}{Scene} & \multirow{2}{*}{Metric} & MRPNet+GS & EDI+GS & EFNet+GS & BAD-NeRF & BAD-GS & E2NeRF & Ev-DeblurNeRF & \ours & \ourspp \\
         & &\cite{zamir2021multi}&\cite{pan2019bringing} &\cite{sun2022event} &\cite{wang2023bad} &\cite{zhao2025bad} & \cite{qi2023e2nerf} & \cite{cannici2024mitigating} & (Ours) & (Ours) \\
         \hline
         \multirow{5}{*}{\textit{factory}}& PSNR$\uparrow$ & 17.44 & 22.46 & 19.74 & 18.81 & 21.35 & 22.28 & 23.33 & \textbf{26.54} & \underline{26.00} \\
         & SSIM$\uparrow$ & 0.5918 & 0.7629 & 0.6415 & 0.6038 & 0.6709 & 0.7822 & 0.8189 & \textbf{0.8856} & \underline{0.8707} \\
         & LPIPS$\downarrow$ & 0.3817 & 0.1448 & 0.3319 & 0.2822 & 0.2391 & 0.1838 & 0.1858 & \textbf{0.0898} & \underline{0.0962} \\
         & MUSIQ$\uparrow$ & 26.18 & \underline{56.74} & 37.12 & 27.43 & 36.19 & 45.88 & 41.58 & 54.24 & \textbf{57.62} \\
         & CLIP-IQA$\uparrow$ & 0.2211 & 0.2177 & 0.2118 & 0.1668 & 0.1718 & 0.2014 & 0.1947 &\underline{0.2215} & \textbf{0.2270} \\
         \hline
         \multirow{5}{*}{\textit{pool}}& PSNR$\uparrow$ & 19.49 & 24.83 & 21.79 & 25.58 & 26.18 & \textbf{27.63} & 27.26 & \underline{27.40} & 26.5880 \\
         & SSIM$\uparrow$ & 0.4718 & 0.6496 & 0.5238 & 0.6888 & 0.7418 & \underline{0.7488} & 0.7440 & \textbf{0.7512} & 0.7283 \\
         & LPIPS$\downarrow$ & 0.3219 & 0.1897 & 0.3718 & 0.2601 & 0.2118 & 0.1995 & 0.2230 & \underline{0.1895} & \textbf{0.1827} \\
         & MUSIQ$\uparrow$ & 15.19 & 47.12 & 26.14 & 30.81 & 39.14 & 47.68 & 44.63 & \underline{51.01} & \textbf{53.03} \\
         & CLIP-IQA$\uparrow$ & 0.1729 & 0.2106 & 0.2037 & 0.1860 & 0.1911 & 0.2473 & \textbf{0.2635} & 0.2126 & \underline{0.2384} \\
         \hline
          \multirow{5}{*}{\textit{tanabata}}& PSNR$\uparrow$ & 18.54 & 23.02 & 20.80 & 16.91 & 20.18 & 23.43 & 23.74 & \textbf{26.18} & \underline{25.90} \\
         & SSIM$\uparrow$ & 0.6203 & 0.8088 & 0.6817 & 0.6483 & 0.7661 & 0.8156 & 0.8059 & \textbf{0.8965} & \underline{0.8896} \\
         & LPIPS$\downarrow$ & 0.3645 & 0.1232 & 0.3128 & 0.2175 & 0.1608 & 0.1505 & 0.1727 & \underline{0.0754} & \textbf{0.0712} \\
         & MUSIQ$\uparrow$ & 28.71 & 59.13 & 39.28 & 17.56 & 27.19 & 47.81 & 41.53 & \underline{63.94} & \textbf{65.54} \\
         & CLIP-IQA$\uparrow$ & 0.2818 & \underline{0.3288} & 0.2518 & 0.2018 & 0.2167 & 0.1914 & 0.2572 & 0.3115 & \textbf{0.3504} \\    
         \hline
          \multirow{5}{*}{\textit{trolley}}& PSNR$\uparrow$ & 19.58 & 24.43 & 21.79 & 17.81 & 21.20 & 24.83 & 24.70 & \textbf{26.67} & \underline{26.43} \\
         & SSIM$\uparrow$ & 0.6811 & 0.8563 & 0.7182 & 0.6114 & 0.7064 & 0.8505 & 0.8465 & \textbf{0.9094} & \underline{0.9026} \\
         & LPIPS$\downarrow$ & 0.3499 & \underline{0.0923} & 0.2691 & 0.2362 & 0.1931 & 0.1157 & 0.1335 & \textbf{0.0708} & \textbf{0.0708} \\
         & MUSIQ$\uparrow$ & 26.39 & 57.56 & 37.98 & 18.71 & 27.19 & 47.87 & 41.80 & \underline{61.48} & \textbf{63.43} \\
         & CLIP-IQA$\uparrow$ & 0.2894 & 0.3434 & 0.2583 & 0.2007 & 0.2176 & 0.2113 & 0.2047 & \underline{0.3618} & \textbf{0.3683} \\
         \bottomrule
    \end{tabular}}
    \newline
    \caption{\textbf{Quantitative comparisons on novel-view synthesis in 4 synthetic scenes}}
    \label{tab:quantitative_synthetic}
\end{table*}

\begin{figure*}[t]
\captionsetup[subfigure]{}
  \centering
  \begin{subfigure}{0.195\linewidth}\includegraphics[width=1.0\linewidth]{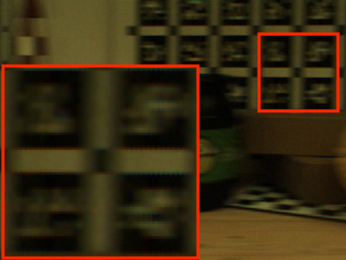}
  \end{subfigure}
  \begin{subfigure}{0.195\linewidth}\includegraphics[width=1.0\linewidth]{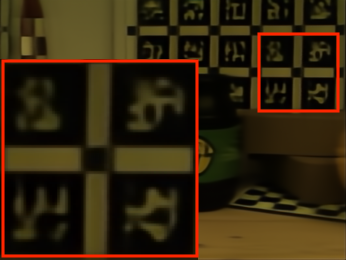}
  \end{subfigure}
  \begin{subfigure}{0.195\linewidth}\includegraphics[width=1.0\linewidth]{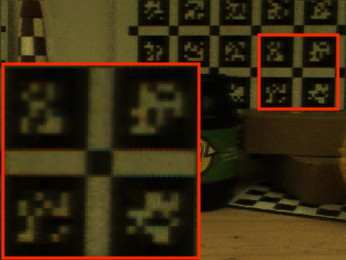}
  \end{subfigure}
  \begin{subfigure}{0.195\linewidth}\includegraphics[width=1.0\linewidth]{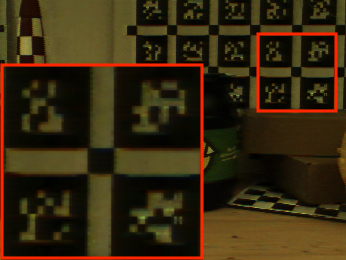}
  \end{subfigure}
  \begin{subfigure}{0.195\linewidth}\includegraphics[width=1.0\linewidth]{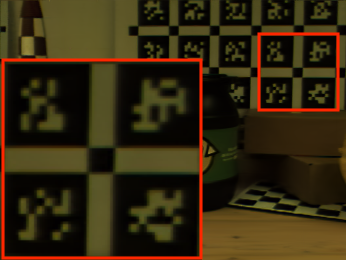}
  \end{subfigure}
  
  \begin{subfigure}{0.195\linewidth}\includegraphics[width=1.0\linewidth]{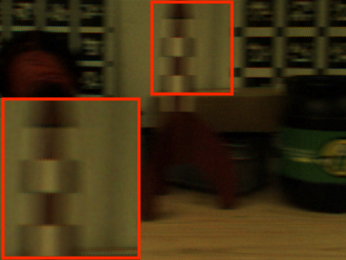}
  \end{subfigure}
  \begin{subfigure}{0.195\linewidth}\includegraphics[width=1.0\linewidth]{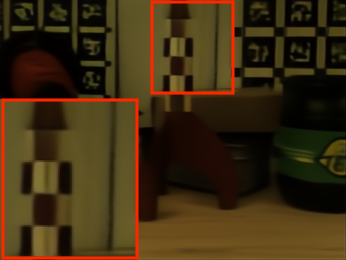}
  \end{subfigure}
  \begin{subfigure}{0.195\linewidth}\includegraphics[width=1.0\linewidth]{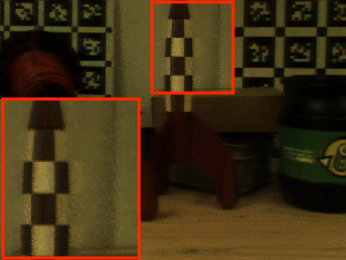}
  \end{subfigure}
  \begin{subfigure}{0.195\linewidth}\includegraphics[width=1.0\linewidth]{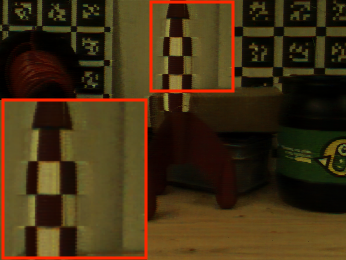}
  \end{subfigure}
  \begin{subfigure}{0.195\linewidth}\includegraphics[width=1.0\linewidth]{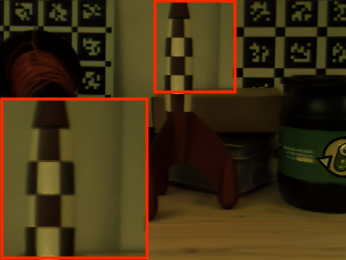}
  \end{subfigure}
  
   \begin{subfigure}{0.195\linewidth}\includegraphics[width=1.0\linewidth]{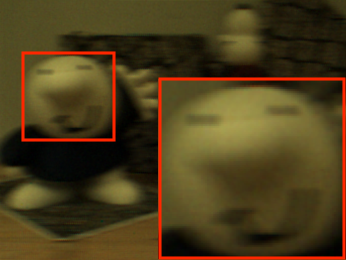}
  \end{subfigure}
  \begin{subfigure}{0.195\linewidth}\includegraphics[width=1.0\linewidth]{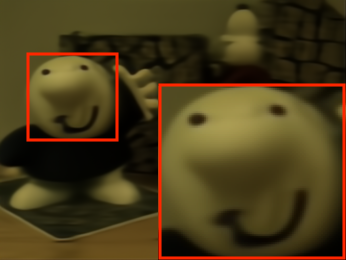}
  \end{subfigure}
  \begin{subfigure}{0.195\linewidth}\includegraphics[width=1.0\linewidth]{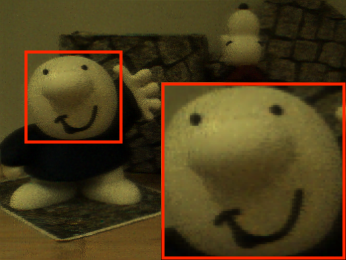}
  \end{subfigure}
  \begin{subfigure}{0.195\linewidth}\includegraphics[width=1.0\linewidth]{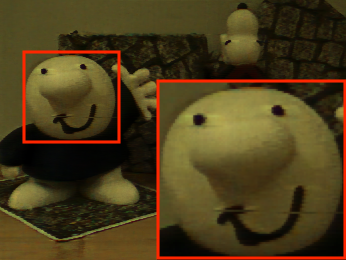}
  \end{subfigure}
  \begin{subfigure}{0.195\linewidth}\includegraphics[width=1.0\linewidth]{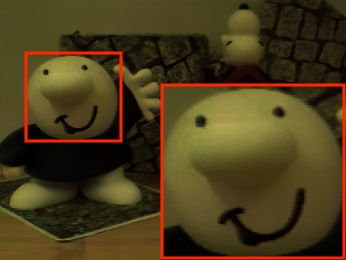}
  \end{subfigure}

  \begin{subfigure}{0.195\linewidth}\includegraphics[width=1.0\linewidth]{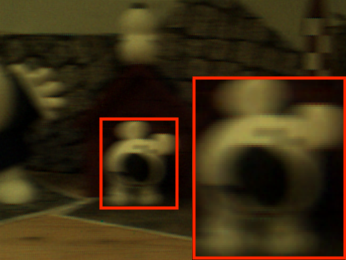}
  \end{subfigure}
  \begin{subfigure}{0.195\linewidth}\includegraphics[width=1.0\linewidth]{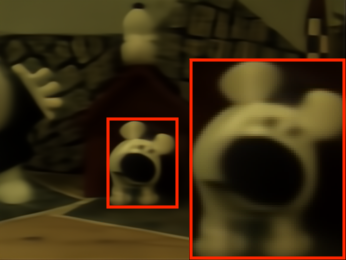}
  \end{subfigure}
  \begin{subfigure}{0.195\linewidth}\includegraphics[width=1.0\linewidth]{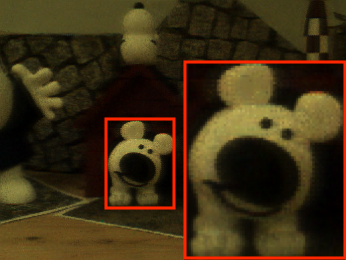}
  \end{subfigure}
  \begin{subfigure}{0.195\linewidth}\includegraphics[width=1.0\linewidth]{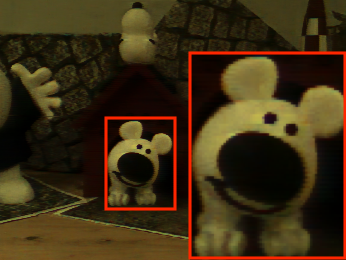}
  \end{subfigure}
  \begin{subfigure}{0.195\linewidth}\includegraphics[width=1.0\linewidth]{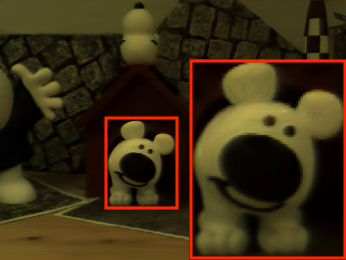}
  \end{subfigure}
  
  \begin{subfigure}{0.195\linewidth}\includegraphics[width=1.0\linewidth]{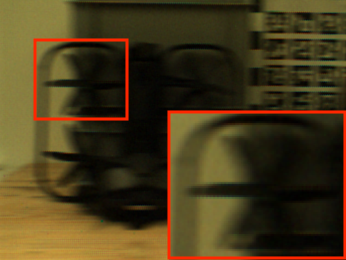}
  \end{subfigure}
  \begin{subfigure}{0.195\linewidth}\includegraphics[width=1.0\linewidth]{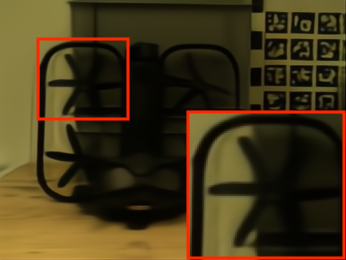}
  \end{subfigure}
  \begin{subfigure}{0.195\linewidth}\includegraphics[width=1.0\linewidth]{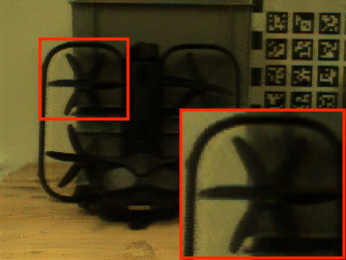}
  \end{subfigure}
  \begin{subfigure}{0.195\linewidth}\includegraphics[width=1.0\linewidth]{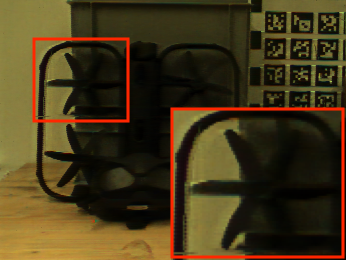}
  \end{subfigure}
  \begin{subfigure}{0.195\linewidth}\includegraphics[width=1.0\linewidth]{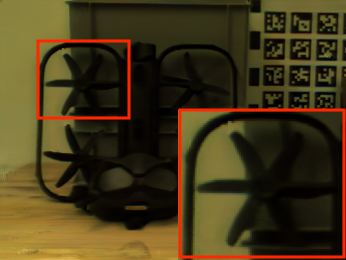}
  \end{subfigure}

  \begin{subfigure}{0.195\linewidth}\includegraphics[width=1.0\linewidth]{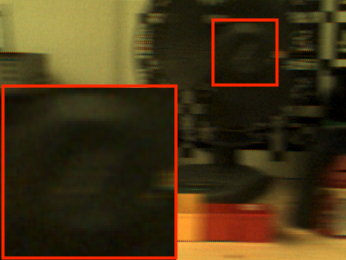}
  \end{subfigure}
  \begin{subfigure}{0.195\linewidth}\includegraphics[width=1.0\linewidth]{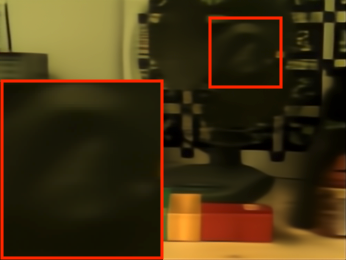}
  \end{subfigure}
  \begin{subfigure}{0.195\linewidth}\includegraphics[width=1.0\linewidth]{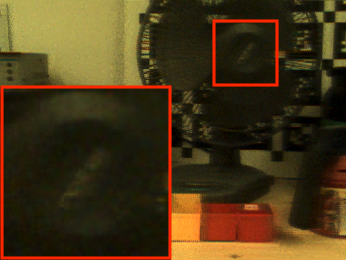}
  \end{subfigure}
  \begin{subfigure}{0.195\linewidth}\includegraphics[width=1.0\linewidth]{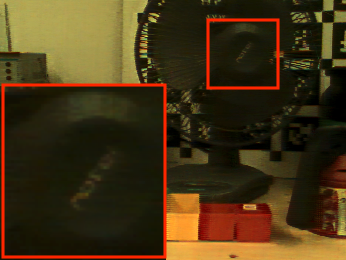}
  \end{subfigure}
  \begin{subfigure}{0.195\linewidth}\includegraphics[width=1.0\linewidth]{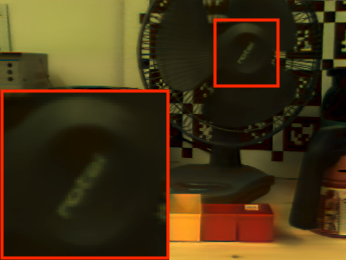}
  \end{subfigure}
  
  \begin{subfigure}{0.195\linewidth}\includegraphics[width=1.0\linewidth]{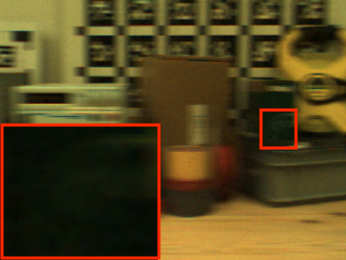}
  \end{subfigure}
  \begin{subfigure}{0.195\linewidth}\includegraphics[width=1.0\linewidth]{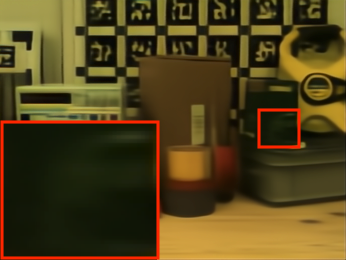}
  \end{subfigure}
  \begin{subfigure}{0.195\linewidth}\includegraphics[width=1.0\linewidth]{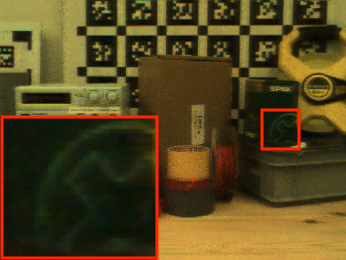}
  \end{subfigure}
  \begin{subfigure}{0.195\linewidth}\includegraphics[width=1.0\linewidth]{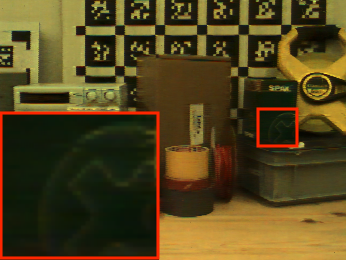}
  \end{subfigure}
  \begin{subfigure}{0.195\linewidth}\includegraphics[width=1.0\linewidth]{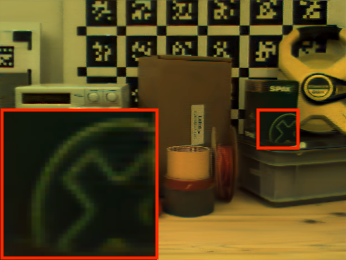}
  \end{subfigure}
  
  \begin{subfigure}{0.195\linewidth}\includegraphics[width=1.0\linewidth]{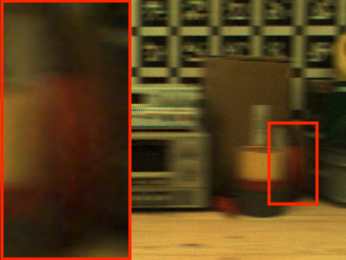}
  \subcaption{\scriptsize{Blur Image}}
  \end{subfigure}
  \begin{subfigure}{0.195\linewidth}\includegraphics[width=1.0\linewidth]{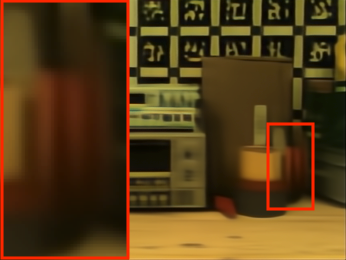}
  \subcaption{\scriptsize{NAFNet~\cite{chen2022simple}}}
  \end{subfigure}
  \begin{subfigure}{0.195\linewidth}\includegraphics[width=1.0\linewidth]{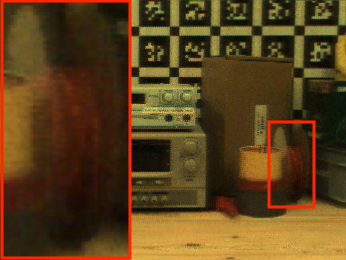}
  \subcaption{\scriptsize{EDI~\cite{pan2019bringing}}}
  \end{subfigure}
  \begin{subfigure}{0.195\linewidth}\includegraphics[width=1.0\linewidth]{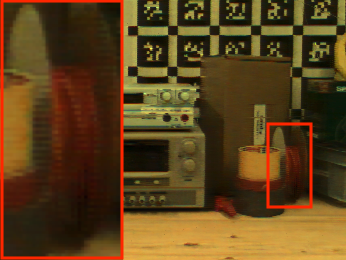}
  \subcaption{\scriptsize{BeNeRF~\cite{li2025benerf}}}
  \end{subfigure}
  \begin{subfigure}{0.195\linewidth}\includegraphics[width=1.0\linewidth]{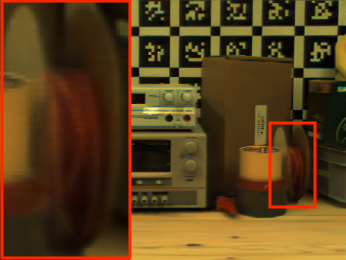}
  \subcaption{\scriptsize{\ourspp (Ours)}}
  \end{subfigure}
  \caption{\textbf{Qualitative comparisons on single image deblurring with real-world datasets.}}
  \label{fig:quaitative_single}
\end{figure*}

\begin{figure*}[t]
\captionsetup[subfigure]{}
  \centering
  \begin{subfigure}{0.195\linewidth}\includegraphics[width=1.0\linewidth]{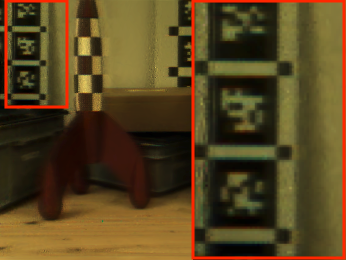}
  \end{subfigure}
  \begin{subfigure}{0.195\linewidth}\includegraphics[width=1.0\linewidth]{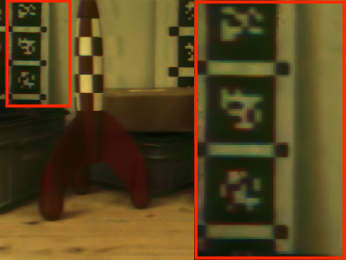}
  \end{subfigure}
  \begin{subfigure}{0.195\linewidth}\includegraphics[width=1.0\linewidth]{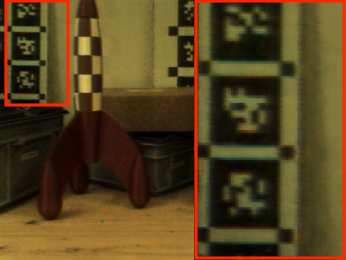}
  \end{subfigure}
  \begin{subfigure}{0.195\linewidth}\includegraphics[width=1.0\linewidth]{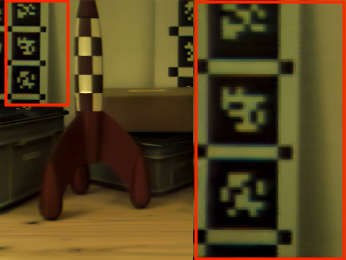}
  \end{subfigure}
  \begin{subfigure}{0.195\linewidth}\includegraphics[width=1.0\linewidth]{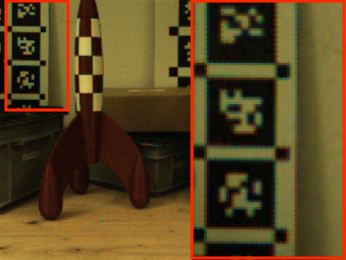}
  \end{subfigure}
  
  \begin{subfigure}{0.195\linewidth}\includegraphics[width=1.0\linewidth]{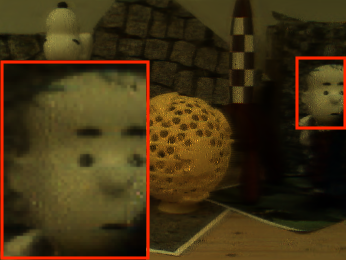}
  \end{subfigure}
  \begin{subfigure}{0.195\linewidth}\includegraphics[width=1.0\linewidth]{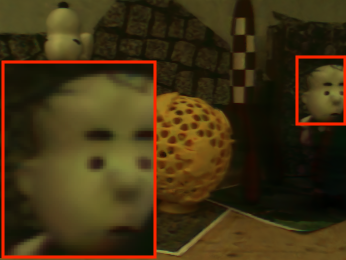}
  \end{subfigure}
  \begin{subfigure}{0.195\linewidth}\includegraphics[width=1.0\linewidth]{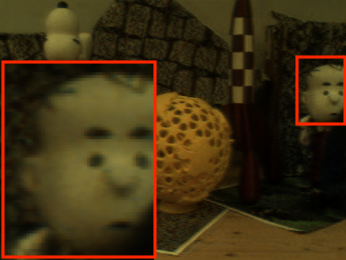}
  \end{subfigure}
  \begin{subfigure}{0.195\linewidth}\includegraphics[width=1.0\linewidth]{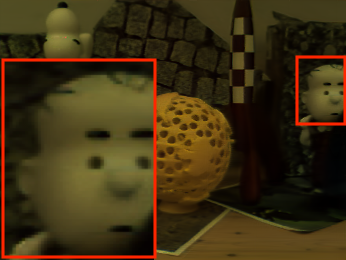}
  \end{subfigure}
  \begin{subfigure}{0.195\linewidth}\includegraphics[width=1.0\linewidth]{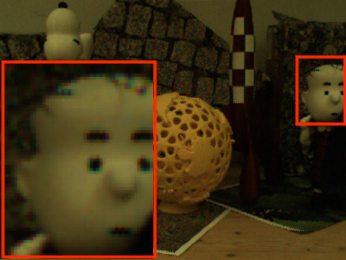}
  \end{subfigure}
  
   \begin{subfigure}{0.195\linewidth}\includegraphics[width=1.0\linewidth]{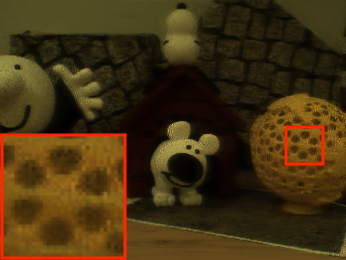}
  \end{subfigure}
  \begin{subfigure}{0.195\linewidth}\includegraphics[width=1.0\linewidth]{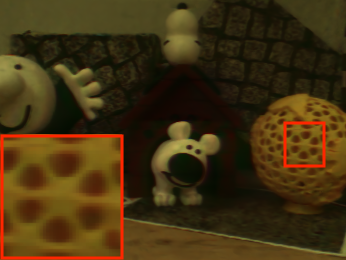}
  \end{subfigure}
  \begin{subfigure}{0.195\linewidth}\includegraphics[width=1.0\linewidth]{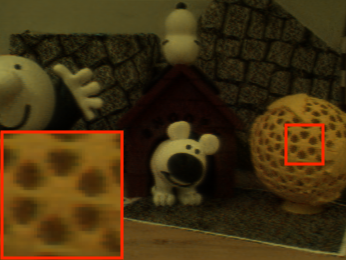}
  \end{subfigure}
  \begin{subfigure}{0.195\linewidth}\includegraphics[width=1.0\linewidth]{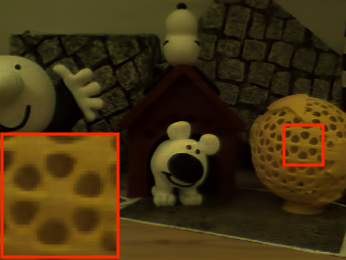}
  \end{subfigure}
  \begin{subfigure}{0.195\linewidth}\includegraphics[width=1.0\linewidth]{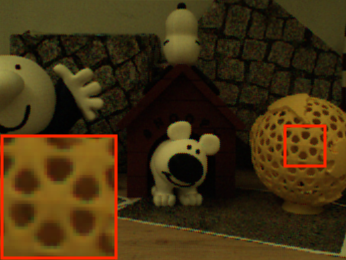}
  \end{subfigure}

  \begin{subfigure}{0.195\linewidth}\includegraphics[width=1.0\linewidth]{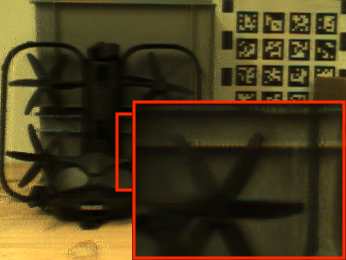}
  \end{subfigure}
  \begin{subfigure}{0.195\linewidth}\includegraphics[width=1.0\linewidth]{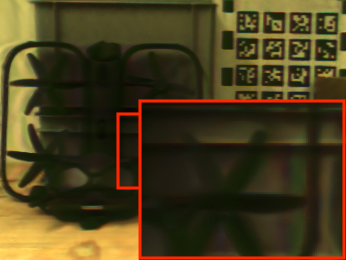}
  \end{subfigure}
  \begin{subfigure}{0.195\linewidth}\includegraphics[width=1.0\linewidth]{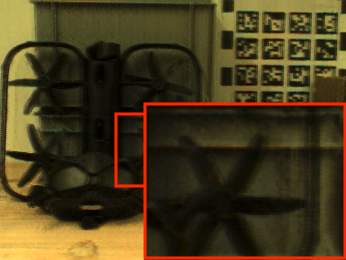}
  \end{subfigure}
  \begin{subfigure}{0.195\linewidth}\includegraphics[width=1.0\linewidth]{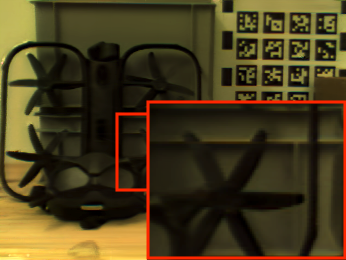}
  \end{subfigure}
  \begin{subfigure}{0.195\linewidth}\includegraphics[width=1.0\linewidth]{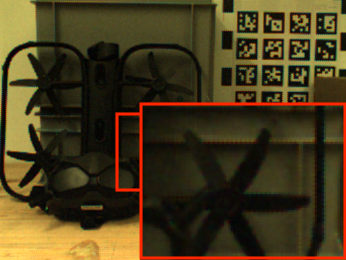}
  \end{subfigure}
  
  \begin{subfigure}{0.195\linewidth}\includegraphics[width=1.0\linewidth]{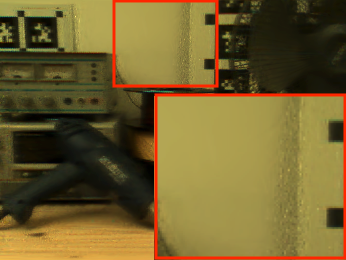}
  \end{subfigure}
  \begin{subfigure}{0.195\linewidth}\includegraphics[width=1.0\linewidth]{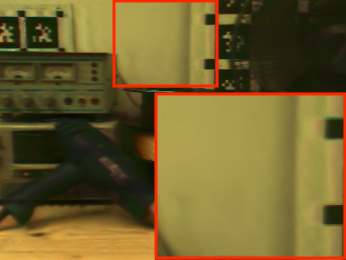}
  \end{subfigure}
  \begin{subfigure}{0.195\linewidth}\includegraphics[width=1.0\linewidth]{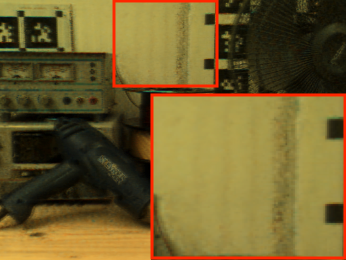}
  \end{subfigure}
  \begin{subfigure}{0.195\linewidth}\includegraphics[width=1.0\linewidth]{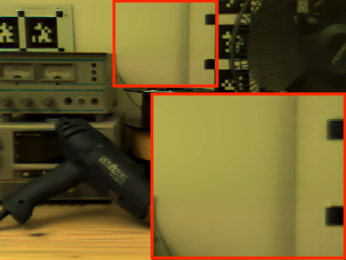}
  \end{subfigure}
  \begin{subfigure}{0.195\linewidth}\includegraphics[width=1.0\linewidth]{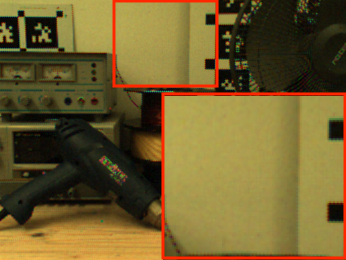}
  \end{subfigure}

  \begin{subfigure}{0.195\linewidth}\includegraphics[width=1.0\linewidth]{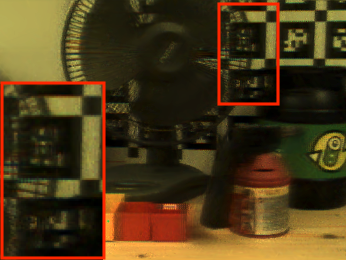}
  \end{subfigure}
  \begin{subfigure}{0.195\linewidth}\includegraphics[width=1.0\linewidth]{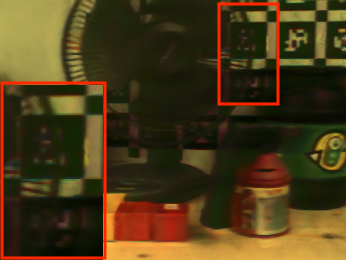}
  \end{subfigure}
  \begin{subfigure}{0.195\linewidth}\includegraphics[width=1.0\linewidth]{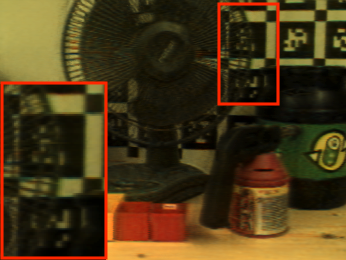}
  \end{subfigure}
  \begin{subfigure}{0.195\linewidth}\includegraphics[width=1.0\linewidth]{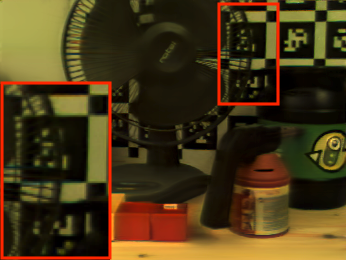}
  \end{subfigure}
  \begin{subfigure}{0.195\linewidth}\includegraphics[width=1.0\linewidth]{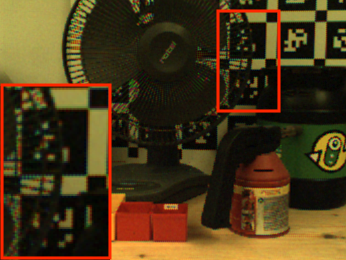}
  \end{subfigure}
  
  \begin{subfigure}{0.195\linewidth}\includegraphics[width=1.0\linewidth]{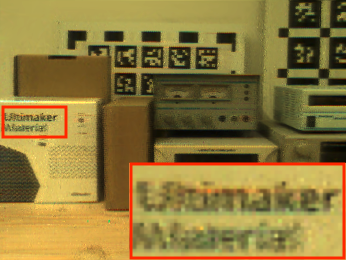}
  \end{subfigure}
  \begin{subfigure}{0.195\linewidth}\includegraphics[width=1.0\linewidth]{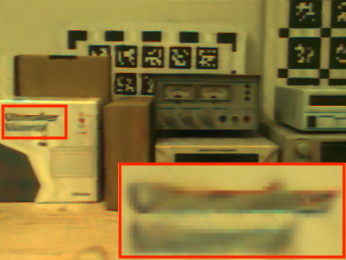}
  \end{subfigure}
  \begin{subfigure}{0.195\linewidth}\includegraphics[width=1.0\linewidth]{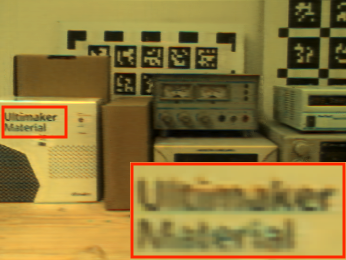}
  \end{subfigure}
  \begin{subfigure}{0.195\linewidth}\includegraphics[width=1.0\linewidth]{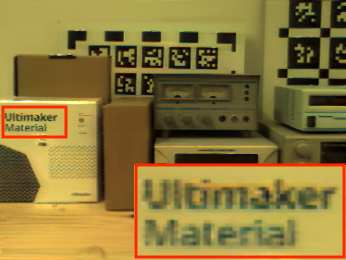}
  \end{subfigure}
  \begin{subfigure}{0.195\linewidth}\includegraphics[width=1.0\linewidth]{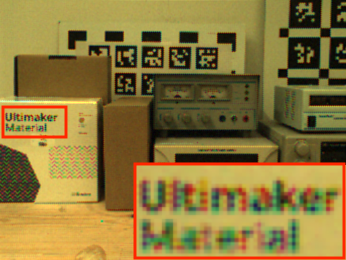}
  \end{subfigure}
  
  \begin{subfigure}{0.195\linewidth}\includegraphics[width=1.0\linewidth]{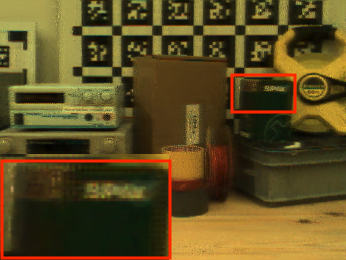}
  \subcaption{\scriptsize{EDI+GS~\cite{pan2019bringing}}}
  \end{subfigure}
  \begin{subfigure}{0.195\linewidth}\includegraphics[width=1.0\linewidth]{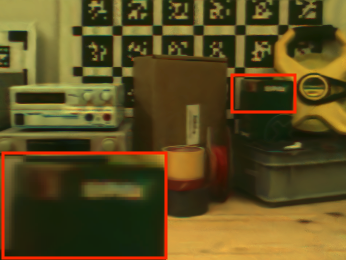}
  \subcaption{\scriptsize{E2NeRF~\cite{qi2023e2nerf}}}
  \end{subfigure}
  \begin{subfigure}{0.195\linewidth}\includegraphics[width=1.0\linewidth]{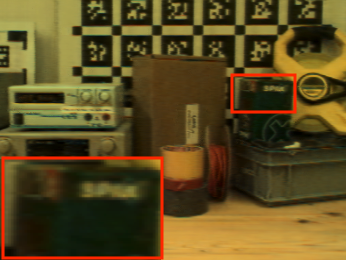}
  \subcaption{\scriptsize{Ev-DeblurNeRF~\cite{cannici2024mitigating}}}
  \end{subfigure}
  \begin{subfigure}{0.195\linewidth}\includegraphics[width=1.0\linewidth]{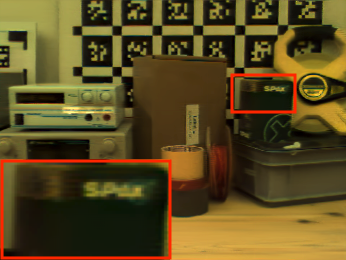}
  \subcaption{\scriptsize{\ourspp (Ours)}}
  \end{subfigure}
  \begin{subfigure}{0.195\linewidth}\includegraphics[width=1.0\linewidth]{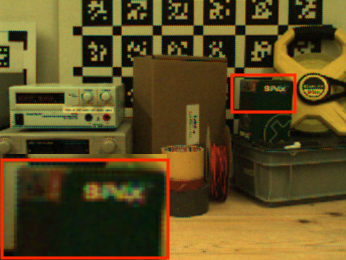}
  \subcaption{\scriptsize{GT}}
  \end{subfigure}
  \caption{\textbf{More qualitative comparisons on novel-view synthesis in real-world datasets.}}
  \label{fig:quaitative_real}
\end{figure*}

\begin{figure*}[t]
\captionsetup[subfigure]{}
  \centering
  \begin{subfigure}{0.195\linewidth}\includegraphics[width=1.0\linewidth]{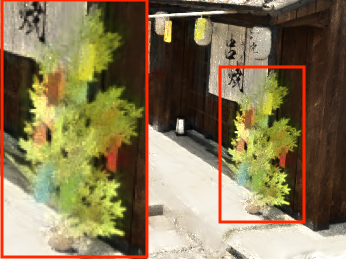}
  \end{subfigure}
  \begin{subfigure}{0.195\linewidth}\includegraphics[width=1.0\linewidth]{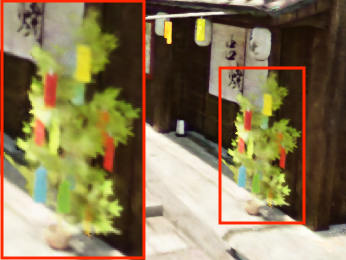}
  \end{subfigure}
  \begin{subfigure}{0.195\linewidth}\includegraphics[width=1.0\linewidth]{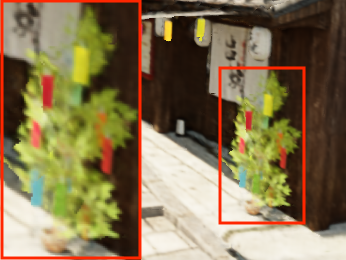}
  \end{subfigure}
  \begin{subfigure}{0.195\linewidth}\includegraphics[width=1.0\linewidth]{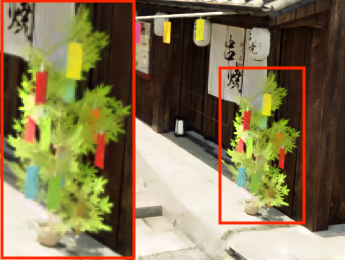}
  \end{subfigure}
  \begin{subfigure}{0.195\linewidth}\includegraphics[width=1.0\linewidth]{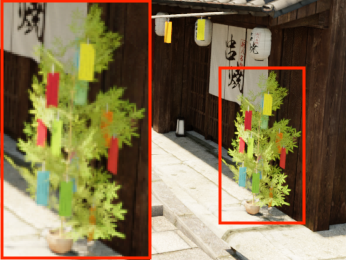}
  \end{subfigure}

  \begin{subfigure}{0.195\linewidth}\includegraphics[width=1.0\linewidth]{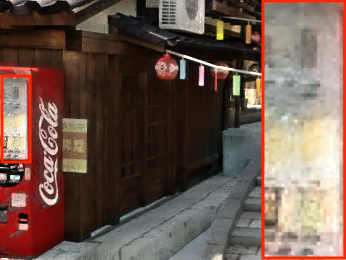}
  \end{subfigure}
  \begin{subfigure}{0.195\linewidth}\includegraphics[width=1.0\linewidth]{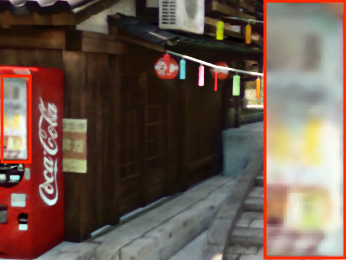}
  \end{subfigure}
  \begin{subfigure}{0.195\linewidth}\includegraphics[width=1.0\linewidth]{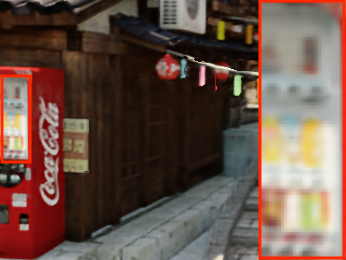}
  \end{subfigure}
  \begin{subfigure}{0.195\linewidth}\includegraphics[width=1.0\linewidth]{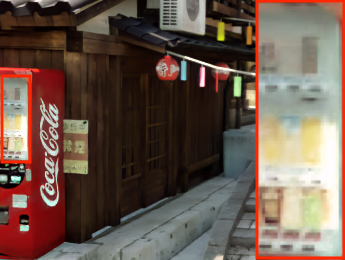}
  \end{subfigure}
  \begin{subfigure}{0.195\linewidth}\includegraphics[width=1.0\linewidth]{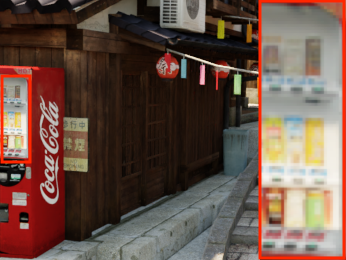}
  \end{subfigure}

  \begin{subfigure}{0.195\linewidth}\includegraphics[width=1.0\linewidth]{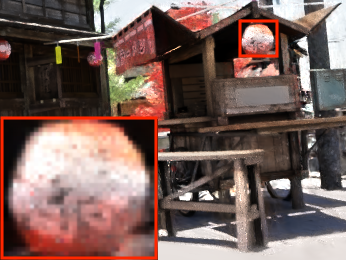}
  \end{subfigure}
  \begin{subfigure}{0.195\linewidth}\includegraphics[width=1.0\linewidth]{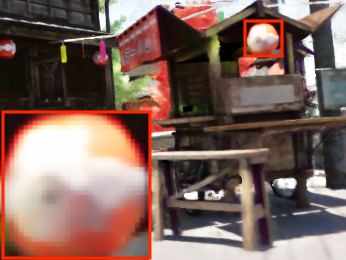}
  \end{subfigure}
  \begin{subfigure}{0.195\linewidth}\includegraphics[width=1.0\linewidth]{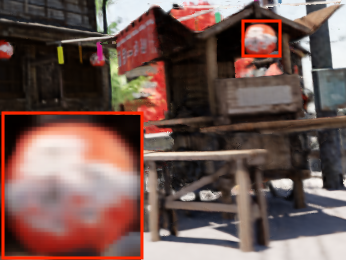}
  \end{subfigure}
  \begin{subfigure}{0.195\linewidth}\includegraphics[width=1.0\linewidth]{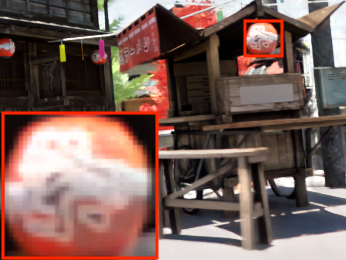}
  \end{subfigure}
  \begin{subfigure}{0.195\linewidth}\includegraphics[width=1.0\linewidth]{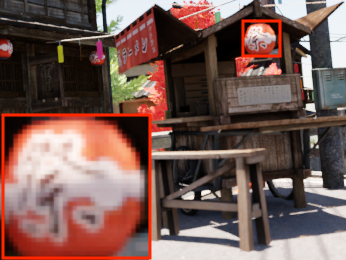}
  \end{subfigure}

  \begin{subfigure}{0.195\linewidth}\includegraphics[width=1.0\linewidth]{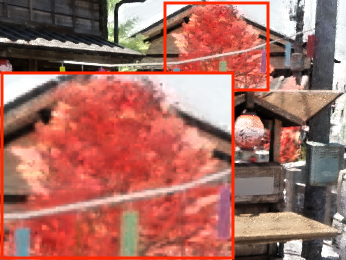}
  \end{subfigure}
  \begin{subfigure}{0.195\linewidth}\includegraphics[width=1.0\linewidth]{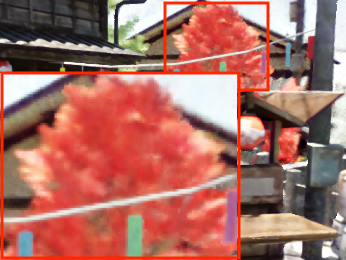}
  \end{subfigure}
  \begin{subfigure}{0.195\linewidth}\includegraphics[width=1.0\linewidth]{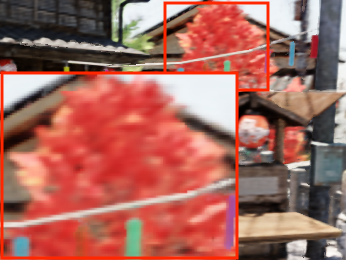}
  \end{subfigure}
  \begin{subfigure}{0.195\linewidth}\includegraphics[width=1.0\linewidth]{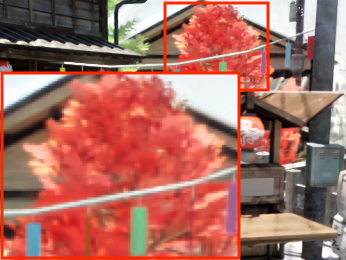}
  \end{subfigure}
  \begin{subfigure}{0.195\linewidth}\includegraphics[width=1.0\linewidth]{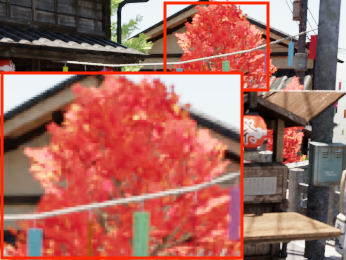}
  \end{subfigure}

   \begin{subfigure}{0.195\linewidth}\includegraphics[width=1.0\linewidth]{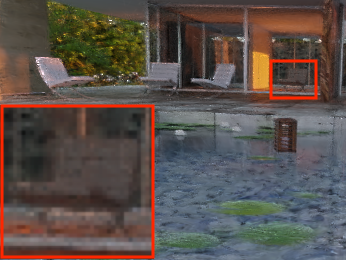}
  \end{subfigure}
  \begin{subfigure}{0.195\linewidth}\includegraphics[width=1.0\linewidth]{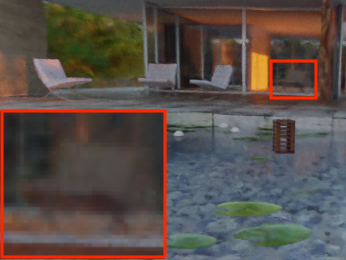}
  \end{subfigure}
  \begin{subfigure}{0.195\linewidth}\includegraphics[width=1.0\linewidth]{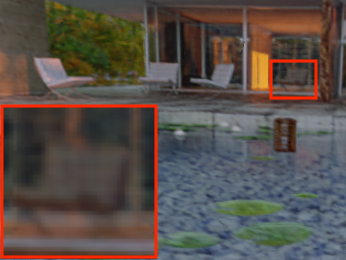}
  \end{subfigure}
  \begin{subfigure}{0.195\linewidth}\includegraphics[width=1.0\linewidth]{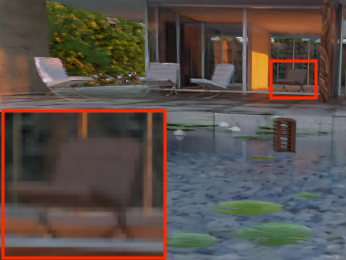}
  \end{subfigure}
  \begin{subfigure}{0.195\linewidth}\includegraphics[width=1.0\linewidth]{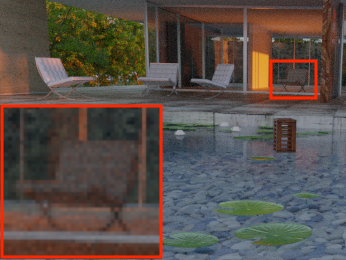}
  \end{subfigure}

  \begin{subfigure}{0.195\linewidth}\includegraphics[width=1.0\linewidth]{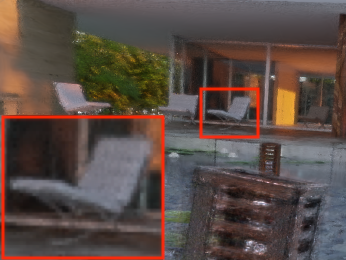}
  \end{subfigure}
  \begin{subfigure}{0.195\linewidth}\includegraphics[width=1.0\linewidth]{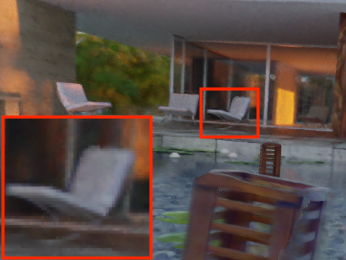}
  \end{subfigure}
  \begin{subfigure}{0.195\linewidth}\includegraphics[width=1.0\linewidth]{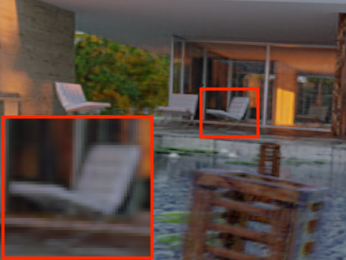}
  \end{subfigure}
  \begin{subfigure}{0.195\linewidth}\includegraphics[width=1.0\linewidth]{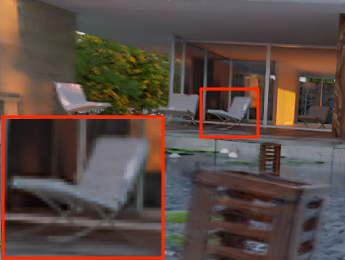}
  \end{subfigure}
  \begin{subfigure}{0.195\linewidth}\includegraphics[width=1.0\linewidth]{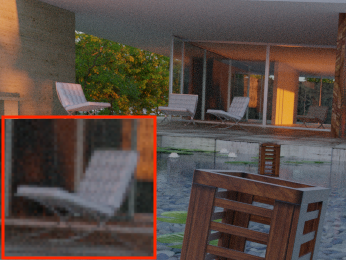}
  \end{subfigure}
  
  \begin{subfigure}{0.195\linewidth}\includegraphics[width=1.0\linewidth]{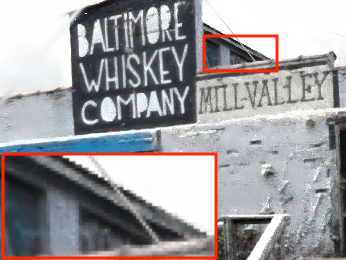}
  \end{subfigure}
  \begin{subfigure}{0.195\linewidth}\includegraphics[width=1.0\linewidth]{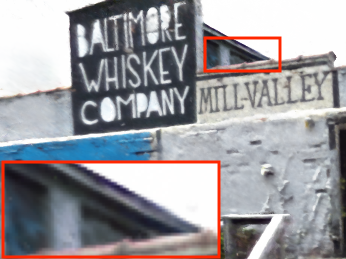}
  \end{subfigure}
  \begin{subfigure}{0.195\linewidth}\includegraphics[width=1.0\linewidth]{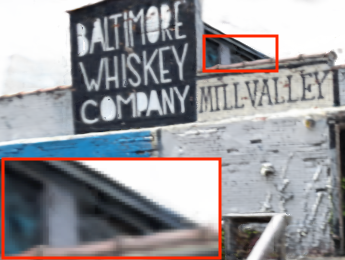}
  \end{subfigure}
  \begin{subfigure}{0.195\linewidth}\includegraphics[width=1.0\linewidth]{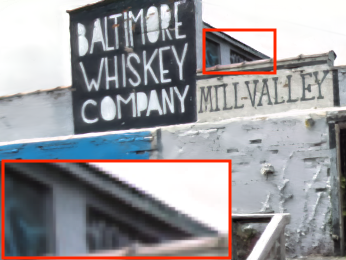}
  \end{subfigure}
  \begin{subfigure}{0.195\linewidth}\includegraphics[width=1.0\linewidth]{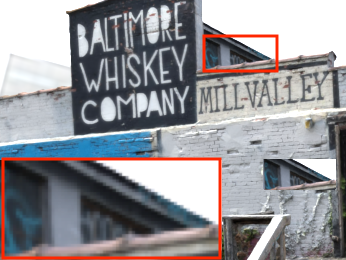}
  \end{subfigure}
  
  \begin{subfigure}{0.195\linewidth}\includegraphics[width=1.0\linewidth]{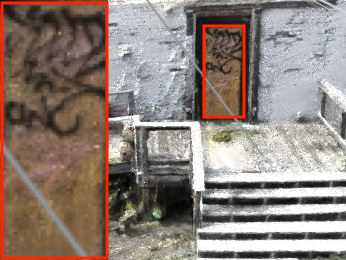}
  \subcaption{\scriptsize{EDI+GS~\cite{pan2019bringing}}}
  \end{subfigure}
  \begin{subfigure}{0.195\linewidth}\includegraphics[width=1.0\linewidth]{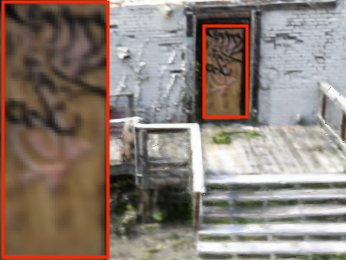}
  \subcaption{\scriptsize{E2NeRF~\cite{qi2023e2nerf}}}
  \end{subfigure}
  \begin{subfigure}{0.195\linewidth}\includegraphics[width=1.0\linewidth]{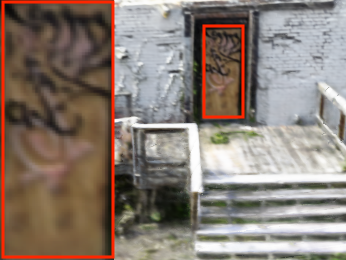}
  \subcaption{\scriptsize{Ev-DeblurNeRF~\cite{cannici2024mitigating}}}
  \end{subfigure}
  \begin{subfigure}{0.195\linewidth}\includegraphics[width=1.0\linewidth]{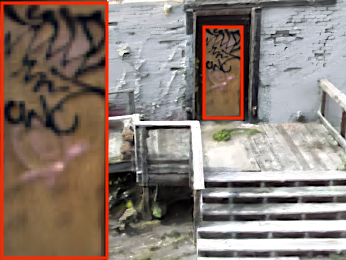}
  \subcaption{\scriptsize{\ourspp (Ours)}}
  \end{subfigure}
  \begin{subfigure}{0.195\linewidth}\includegraphics[width=1.0\linewidth]{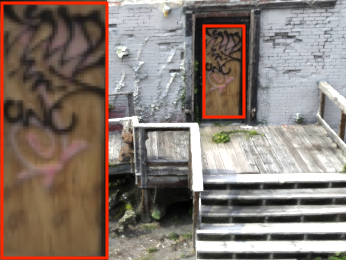}
  \subcaption{\scriptsize{GT}}
  \end{subfigure}
  \caption{\textbf{More qualitative comparisons on novel-view synthesis in synthetic datasets.}}
  \label{fig:qualitative_synthetic}
\end{figure*}

\end{document}